\documentclass{article}
\usepackage{graphicx}
\usepackage{multicol}
\usepackage{microtype}
\usepackage{graphicx}
\usepackage{subfigure}
\usepackage{booktabs}
\usepackage{hyperref}

\usepackage[accepted]{icml2024}
\usepackage{amsmath}
\usepackage{amssymb}
\usepackage{mathtools}
\usepackage{amsthm}
\usepackage{multirow}

\begin{document}

\twocolumn[
\icmltitle{SHYI: Action Support for Contrastive Learning
in High-Fidelity Text-to-Image Generation}
\icmlsetsymbol{equal}{*}
\begin{icmlauthorlist}
\vskip -0.15in
\icmlauthor{Tianxiang Xia\textsuperscript{\textdagger}}{equal}
\icmlauthor{Lin Xiao}{equal}
\icmlauthor{Yannick Montorfani}{equal}
\icmlauthor{Francesco Pavia}{equal}
\icmlauthor{Enis Simsar}{}
\icmlauthor{Thomas Hofmann}{}
\end{icmlauthorlist}
\vskip 0.05in
\hspace{3cm}\textsuperscript{\textdagger}\footnotesize{Email: xiatia@ethz.ch}
]

\section{Introduction}

Diffusion-based text-to-image generators have become increasingly popular in recent years due to their ease of training, leveraging simple and efficient loss functions, and their enhanced stability during training. Despite these advantages, they still face challenges in generating images that faithfully adhere to textual prompts. Common issues include miscounting, missing objects, and errors in attribute binding.

To address these limitations, CONFORM \cite{CONFORM} employs Contrastive Learning as a solution. However, problems persist (interactions are missing or misinterpreted) when it comes to generating images involving interactions between objects
(see Fig.\ \ref{fig:cat_dog} / \ref{fig:cat_mouse} / \ref{fig:cat_dog_obj} / \ref{fig:cat_mouse_obj}).

In this project, we aim to tackle the issue of infidelity in text-to-image generation, focusing particularly on actions involving multiple objects. Building upon the foundations of CONFORM \cite{CONFORM} and InteractDiffusion \cite{InteractiveDiffusion}, our primary goal is to generalize the contrastive learning approach in CONFORM \cite{CONFORM} to better support actions and potentially other complex interactions. Furthermore, we explore the integration of concepts from InteractDiffusion \cite{InteractiveDiffusion} to enhance interaction embeddings within the model, promoting more accurate and contextually appropriate visualizations of object interactions.

In the end, our method shows promising results even with verbs that Stable Diffusion \cite{stable_diffusion} understands mediocrely (c.f.\ \ref{subsec:complete_eval} for a thorough analysis).

\section{Model and Methods}
\label{sec:method}

We publish our codebase on polybox\footnote{\url{https://polybox.ethz.ch/index.php/s/dJm3SWyRohUrFxn} see README.md inside as entrypoint}, which also includes  different implementations and test results.

To simplify the writing, we use acronyms to address our methods: \textbf{S}emantically \textbf{hy}pergraphic contrastive adjacency learning
with \textbf{I}nteractDiffusion enhancement gives the
acronym of our main method: \textbf{SHYI}; And we call our method without InteractDiffusion \textbf{SHY}.

\subsection{Semantically hypergraphic contrastive adjacency learning}

\label{subsec:shy}

\begin{figure}[h]
    \center
    \includegraphics[height=2.5cm]{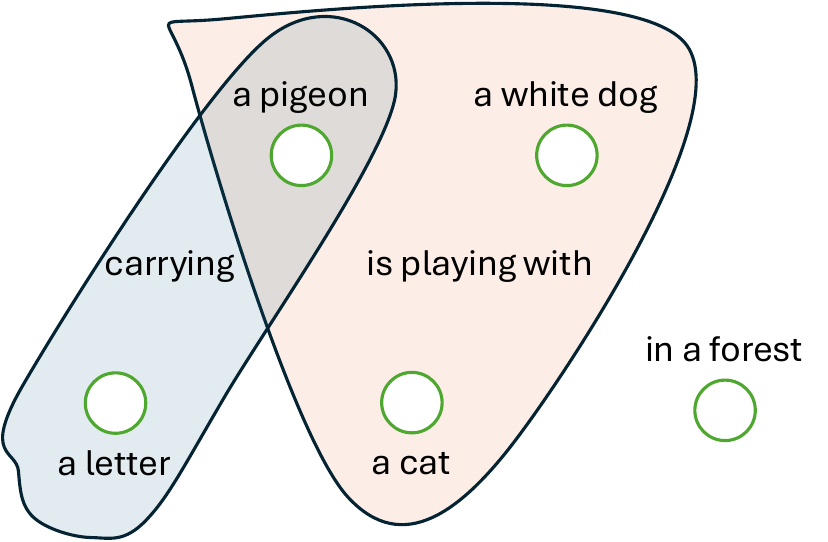}
    \caption{Semantic hypergraph of ``A pigeon carrying a letter is playing with a white dog and a cat in a forest''}
    \label{fig:topo}
\end{figure}
A {\bf Semantic Hypergraph} is constructed from a prompt: noun groups as vertices $V$,
and hyperedges $E = \{a \text{ action of } m \text{ parties } v_1, \ldots, v_m\in V\}$, as in Fig.\ \ref{fig:topo}.

Inspired by \citet{CONFORM}, we first perform {\bf contrastive adjacency learning}
on all vertices to reduce the overlapping of their attention regions.

For each hyperedge ($a$, $v_1, \ldots, v_m$), we perform {\bf contrastive adjacency learning}: we pose a loss on the attention so that the vertices have different attention regions, but the region borders are close to each other.

In addition, we apply {\bf another contrastive learning} between the action with its associated parties and the rest of the components. 
Note that in practice, for this contrastive learning,
we need to use detached PyTorch tensor values of the objects.
This results in zero gradient on the latent variables of the objects on this learning part,
so that the action is drawn to its associated objects
and not the opposite.

One simplistic visual example to demonstrate our method is in Figure \ref{fig:minimal} and a more detailed description of the loss function is in the Appendix \ref{subsec:poloss}.

\begin{figure}[h]
    \center
    \includegraphics[height=2.5cm]{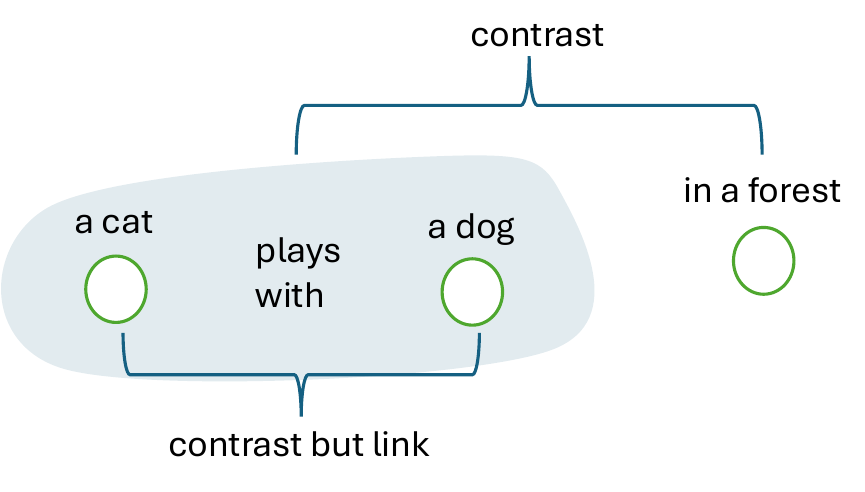}
    \includegraphics[width=0.22\linewidth]{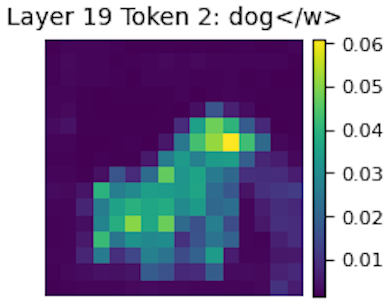}
    \includegraphics[width=0.22\linewidth]{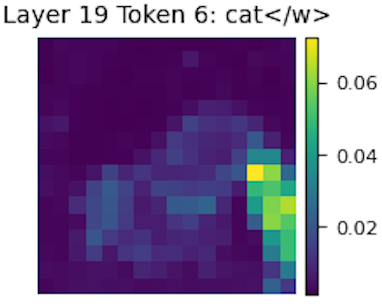}
    \includegraphics[width=0.22\linewidth]{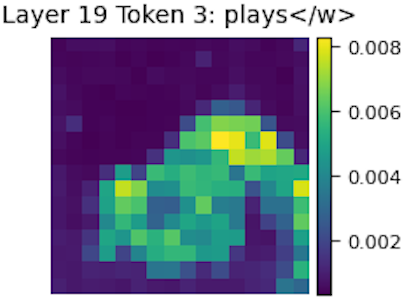}
    \includegraphics[width=0.22\linewidth]{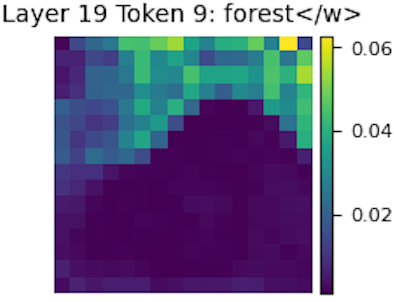}
    \caption{Contrast means that the attention regions are separated and link means that the regions are close; the diagram shows the concept and the plots
    show an example (from SHYI method) of attention maps of tokens where ``dog"
    and ``cat" have distinct yet connected attention regions, and
    the attention region ``plays" is forced onto ``dog" and ``cat" by contrasting with ``forest"}
    \label{fig:minimal}
\end{figure}

\textbf{Iterative Loss Selection} Since this is a multiobjective optimization,
simply adding the losses together
and optimizing by gradient descent does not
perform well. To mitigate this effect,
we alternatively select different loss terms of different
objectives at each
gradient descent iteration.

When a constrastive group has only one token, it pairs
to no one and produces no loss for the network to learn.
To produce a loss value for this token and also to enhance consistency
across timesteps,
we add the attention maps of the same tokens of step $t+1$
into their contrastive groups. This is performed by \citet{CONFORM}
as well.

Finally, the latent representation is updated at each iteration as
$
z_t^{\prime}=z_t-\alpha_t \nabla_{z_t} \mathcal{L}
$
for the selected loss $\mathcal{L}$.

\subsection{Enhancement by InteractDiffusion}

Diffusion models often struggle with accurately generating images from prompts involving  interactions between multiple objects. These challenges primarily come from the models’ inability to properly understand relationships such as subject-action-object and maintain spatial fidelity in intricate scenes. To address these issues, we integrated components of InteractDiffusion \cite{InteractiveDiffusion} into our framework to enhance action understanding.

\paragraph{Leveraging InteractDiffusion Weights}
InteractDiffusion \cite{InteractiveDiffusion} incorporates object interaction-aware attention mechanisms and role-specific embeddings (e.g., subject, object, action) into its fine-tuning process. To benefit from that, we imported the \textbf{pre-trained weights} of InteractDiffusion, which were trained to improve modeling of object interactions. 

While InteractDiffusion typically includes bounding box annotations to guide the model during training and inference, we decided not to include it as our method requires
only on the prompts itself. Instead, the improvements achieved (see Figure \ref{fig:iduseful} and Section \ref{sec:res} in general) came solely from the interaction-specific prior knowledge embedded in the weights.

\paragraph{Integration into the pipeline}
In practice, to integrate InteractDiffusion, we replaced the U-Net\footnote{\url{https://huggingface.co/interactdiffusion/diffusers-v1-2/tree/main/unet}} component in the pipeline with that of InteractDiffusion. Then we made adjustments in the codebase
to make it compatible with this new U-Net architecture and included other components
of the InteractDiffusion model with which the U-Net was trained (although those components
were not trained by the InteractDiffuion team).

\subsection{Related Work}

Previous work has already addressed the infidelity of text-to-image generation for multiple objects with different attributes through contrastive learning (CONFORM \cite{CONFORM}) or segregation loss (A-star \cite{A-star}), ensuring attribute binding consistency (Divide \& Bind \cite{Divide_Bind}) and maximizing attention for all objects in prompts (Attend-and-Excite \cite{Attend-and-Excite}).

These methods typically work on top of diffusion-based text-to-image generation models like Stable Diffusion \cite{StableDiffusion} or Imagen \cite{Imagen}.
InteractDiffusion \cite{InteractiveDiffusion} improves text-to-image generation on prompts that include interaction with multiple objects but relies on region marking. 

Although these methods are good at handling separate objects and their attributes, they face challenges when it comes to dealing with interactions as they cannot be assigned simply to one object.

\section{Results}
\label{sec:res}

Due to the limitation of the relatively small diffusion
models that we use, image generation with
prompts with more than two interacting objects
is catastrophic for all the pipelines. Hence, we stick to
prompts of two interacting objects.

We evaluate our methods through qualitative and quantitative experiments.
A user study is conducted to compare image generation methods (details in \ref{subsec:complete_eval}).

Quantitatively, we use TIFA \cite{TIFA}, which performs text-to-image faithfulness evaluations with question answering. The  question-answer pairs for the TIFA scores are generated using GPT-3.5turbo and GPT-4o. In addition, we evaluate our results using CLIP \cite{clip}: compare the cosine similarity of text and image embeddings.

\subsection{First implementations and tests}
\label{subsec:firsts}

We represent here results from our first incomplete implementations of our method
to justify the different complements of the pipeline.

We first implemented our method without iterative loss selection, but added all the losses together. We also tested
the method with or without the enhancement of InteractDiffusion.

\begin{figure}
    \centering
    \includegraphics[width=0.49\linewidth]{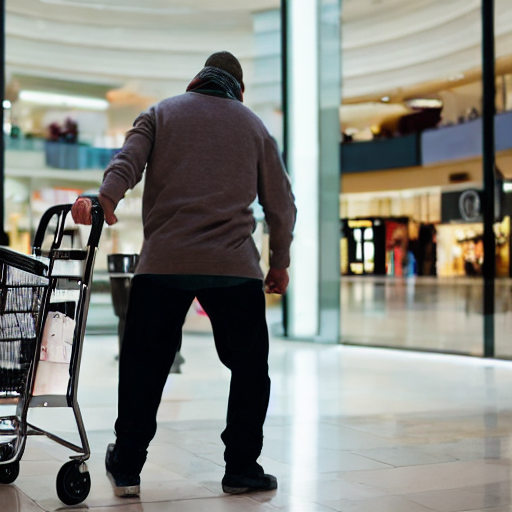}
    \includegraphics[width=0.49\linewidth]{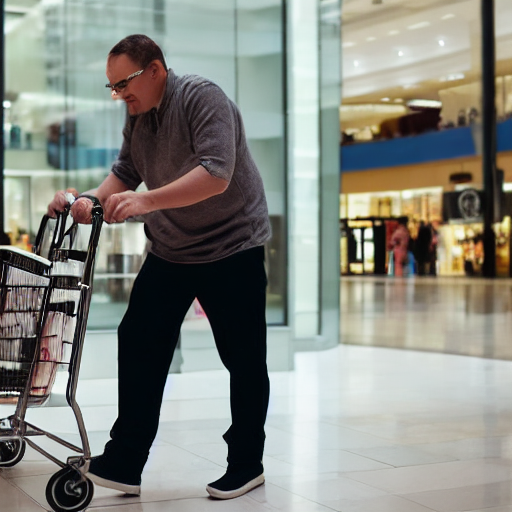}
    \caption{On the left is without InteractDiffusion and on the right is with it;
    the prompt is ``A man pushing a cart in the mall"}
    \label{fig:iduseful}
\end{figure}

In most of the instances, the results of with or without InteractDiffusion
are very similar. However, in one instance, we noticed an evident improvement in fidelity in generating the ``push" action (c.f.\ Figure \ref{fig:iduseful}).

\begin{figure}
    \centering
    \includegraphics[width=0.49\linewidth]{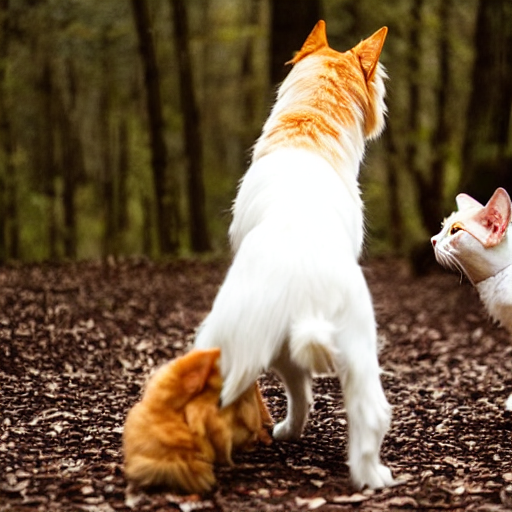}
    \includegraphics[width=0.49\linewidth]{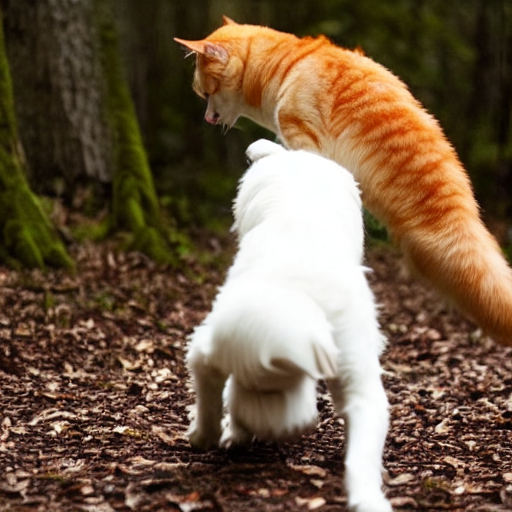}
    \caption{SHYI without (left) or with (right) iterative loss selection;
    the prompt is ``A white dog plays with a ginger cat in a forest"}
    \label{fig:bad}
\end{figure}

The results of SHY and SHYI were poor for the prompt ``A white dog plays with a ginger cat in a forest". The main difference between this prompt
and other prompts was that here we have adjectives in addition, giving more objectives for our method to optimize. This was where we suspected that multiple objectives are the cause of bad quality.
Indeed, the image qualities improved after applying
iterative loss selection, see Figure \ref{fig:bad}.

We observe that our objective of bring interacting objects together (i.e.\ link loss in Figure \ref{fig:minimal}) is mostly achieved.

\begin{figure}
    \centering
    \includegraphics[width=0.32\linewidth]{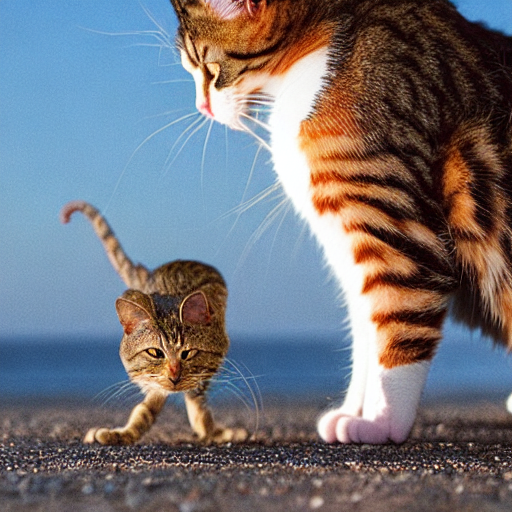}
    \includegraphics[width=0.32\linewidth]{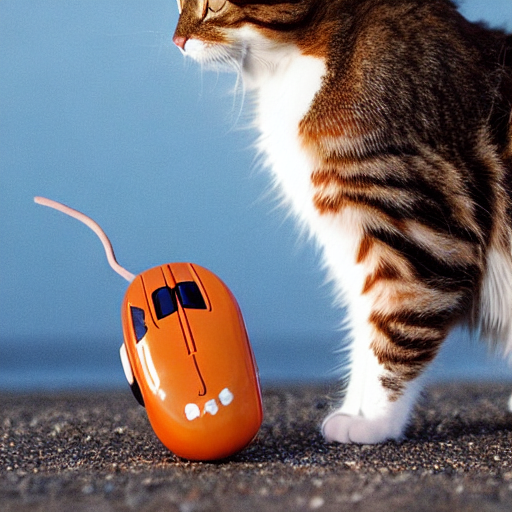}
    \includegraphics[width=0.32\linewidth]{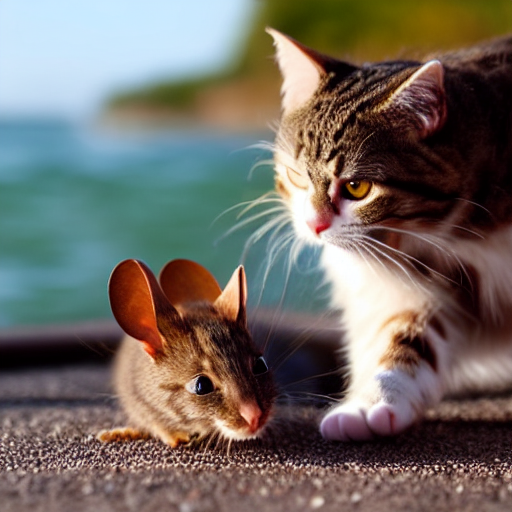}
    \caption{On the left, Stable Diffusion mixed features of objects; on the middle, CONFORM misinterpreted ``mouse"; on the right is the image generated by SHYI where the cat has more corresponding action; the prompt is
    ``A cat catches a mouse by the sea"; they are generated with the same
    seed for the diffusion model thus the positionings are roughly the same}
    \label{fig:others}
\end{figure}

The standard Stable Diffusion\footnote{model from \url{https://huggingface.co/stable-diffusion-v1-5/stable-diffusion-v1-5}} often mixed up features of different
objects; CONFORM sometimes misinterpreted objects, perhaps because it separated
object attention regions too much (c.f.\ Figure \ref{fig:others}).

TIFA score presents some issues when evaluating action fidelity,
indeed the TIFA score was 1 for both images in Figure \ref{fig:iduseful}: it cannot differentiate ``pull" and ``push". Actually the TIFA score was 1 for all pipelines of the prompt ``A man pushing a cart in the mall" in spite of evidential differences in their quality. This leads to the user study for our complete implementation.
As for CLIP score, it is around 30 for all the evaluations, which is not very differentiating in our case.

\subsection{Complete implementation and tests}
\label{subsec:complete_eval}

The experiments in Subsection \ref{subsec:firsts} lead to our implementation of all the components mentioned in Section \ref{sec:method}.

We focus our further testing on CONFORM
and SHYI where CONFORM\footnote{implementation at \url{https://github.com/gemlab-vt/CONFORM}} is our baseline model.

\textbf{Dataset} To not include our bias in designing the tests, we obtain the test dataset by prompting ChatGPT-4-turbo. The exact conversation is included in Appendix \ref{subsec:conversation}.

We first evaluated all the output with TIFA score and CLIP score.
Here the question-answer pairs are generated by GPT-4o.
CONFORM's output achieved 0.89 in average while SHYI's output achieved 0.86.
As for CLIP scores, CONFORM's output achieved 0.315 and SHYI's output achieved 0.312.
Indeed CONFORM's output images have slightly better clarity in some cases (c.f.\ Section \ref{sec:diss}).

\subsection{User Study}
\label{subsec:study}

To comprehensively evaluate the performance of our method, we conducted a user study focusing on qualitative assessments of image generation.

The primary aim of this study was to address the limitations of automated metrics such as TIFA and CLIP, which fail to reliably capture action fidelity and object interactions, as noted in Subsection \ref{subsec:firsts}.

Participants were asked to compare images generated by our pipeline (SHYI) with those generated by the baseline model (CONFORM). They evaluated the images based on:
\begin{enumerate}
    \item Prompt Accuracy: Which image encapsulates the prompt better? 
    \item Interaction Quality: Which image shows more interaction between the two objects?
\end{enumerate}
An example of this evaluation is shown in Fig.\ \ref{fig:google_form}.

The study included around 30 participants.
In total, each participant reviewed 
$
6 \text{ prompts} \times 3 \text{ seeds} \times 2 \text{ questions} = 36 \text{ comparisons}
$.

The study was designed to ensure balanced and comprehensive feedback across a diverse set of prompts. The prompts are selected by first eliminating the prompts that have poor image quality for both of the pipelines and then by choosing randomly. We do this selection to keep the survey form short for participants.

The results for each prompt and each model, averaged across the seeds and questions, are summarized in Table \ref{tab:conform_shyi_table}:

\begin{table}[h!]
    \centering
    \begin{tabular}{|p{4cm}|p{2cm}|}
        \hline
        \multicolumn{2}{|l|}{\textbf{Prompt} Young girl reading a thick book by the window} \\
        \hline
        \textbf{CONFORM} 20.67\% & \textbf{SHYI} 79.33\% \\
        \hline
        \multicolumn{2}{|l|}{\textbf{Prompt} A dog carrying a stick over a bridge} \\
        \hline
        \textbf{CONFORM} 46.58\% & \textbf{SHYI} 53.42\% \\
        \hline
        \multicolumn{2}{|l|}{\textbf{Prompt} Fisherman catching a fish near a river} \\
        \hline
        \textbf{CONFORM} 22.7\% & \textbf{SHYI} 77.3\% \\
        \hline
        \multicolumn{2}{|l|}{\textbf{Prompt} Golden Retriever holding a stick on a beach} \\
        \hline
        \textbf{CONFORM} 58.37\% & \textbf{SHYI} 41.63\% \\
        \hline
        \multicolumn{2}{|l|}{\textbf{Prompt} Red Robot fixing a broken car in a garage} \\
        \hline
        \textbf{CONFORM} 64.23\% & \textbf{SHYI} 35.77\% \\
        \hline
        \multicolumn{2}{|l|}{\textbf{Prompt} Skilled artist carving marble statue in a workshop} \\
        \hline
        \textbf{CONFORM} 61.87\% & \textbf{SHYI} 38.13\% \\
        \hline
        \hline
        \textbf{AVG} 45.73\% & \textbf{AVG} 54.26\% \\
        \hline
    \end{tabular}
    \caption{User study vote percentages averaged across seeds and questions}
    \label{tab:conform_shyi_table}
\end{table}

\textbf{Overall Results}

When averaged across all prompts, SHYI achieved \( 54.16\% \) on Question 1 and \( 54.36\% \) on Question 2, outperforming CONFORM's \( 45.83\% \) on Question 1 and \( 45.63\% \) on Question 2. A collection of test results is in \ref{subsec:collection}.


\section{Discussion}
\label{sec:diss}

We find that the results depend very much on the ability of Stable Diffusion (addressed as SD in the following) to understanding the action verbs.

When SD understands the verb very well, CONFORM and SHYI have comparable fidelity (see Figure \ref{tab:Young girl reading thick book by window}/\ref{tab:Red robot fixing broken car in garage}/\ref{tab:Child reading book by fireplace}).

SD tends to interpret actions in the sense of human being. In this case, SHYI further fosters this effect. Effectively, in Table \ref{tab:Dog carrying stick over bridge} (``Dog carrying stick over bridge"), the dogs are carrying sticks by hand as humans do. In this sense, SHYI outperforms CONFORM. This problem may be solved by training the diffusion model with more animal interactions. When the action is indeed done by human (and when Stable Diffusion interprets this verb not poorly), SHYI clearly outperforms CONFORM as in Table \ref{tab:Fisherman catching fish near river} (``Fisherman catching fish near river").

When SD does not understand the action (or does not interpret it as a verb) at all, both pipelines perform poorly. For example, in Table \ref{tab:Robot painting canvas in studio} (``Robot painting canvas in studio"), SD does not understand ``painting" as a verb, perhaps due to its strong prior of it being a noun. Still, one of SHYI's result (seed 4914) shows a brush held by the robot (in the painting though).

In one case CONFORM clearly outperforms SHYI, see Table \ref{tab:Bird building nest on cliff} (``Bird building nest on cliff"). We think that this is because our link loss is too strong in modifying the latent variables, so that it maximizes the border connection between the two objects. This should be mitigated by decreasing the weight of link loss or making it less often to be selected (in iterative loss selection c.f.\ \ref{subsec:shy}).

Of course, apart from the understanding of actions of SD, other coefficients also influence image quality and fidelity. In Table \ref{tab:Cat chasing mouse under tree} (``Cat chasing mouse under tree"), SHYI shows fushioned cats and mouses. The reason is that the environment (``tree") has too large attention region so ``cat" and ``mouse" are forced to have the same attention region. Nevertheless, SHYI with seed 4914 has the only image output where the cat is chasing.

\section{Summary}

In general, our method SHYI shows promising yet not perfect results in high-fidelity text-to-image generation with object interactions by aiding the attention assignment with semantically hypergraphic contrastive adjacency learning.

To further improve our method, we list here some future directions:
\begin{itemize}
    \item \textbf{Hyperparameter tuning}: to scale to more complexe prompts, our method naturally introduces more losses. So how to balance those losses should be studied and tested.
    \item \textbf{Spatial relationship of objects}: how to induce the correct spatial relationship of objects without manually entering the boxes as in InteractDiffusion \cite{InteractiveDiffusion} to replace the link loss.
    \item \textbf{Improve the understanding of actions} of Stable Diffusion:
    for example, how to force Stable Diffusion to understand ``paint" as a verb instead of noun.
\end{itemize}

\bibliography{bibliography}
\bibliographystyle{icml2024}

\appendix
\section{Appendix}

\subsection{Loss function formulas}
\label{subsec:poloss}

We define:
$$
\begin{aligned}
\mathcal{L}_\text{Contrast}=-\log \frac{\exp \left(\operatorname{sim}\left(A^j, A^{j^{+}}\right) / \tau_1\right)}{\sum_{n \in\{j^{+}, j_1^{-}, \ldots j_N^{-}\}} \exp \left(\operatorname{sim}\left(A^j, A^n\right) / \tau_1\right)}\\
\mathcal{L}_\text{Link}=-\log \frac{\exp \left(\operatorname{sim}\left(\tilde A^j, \tilde A^{j^{-}}\right) / \tau_2 \right)}{\sum_{n \in\{j^{-}, j_1^{+}, \ldots j_N^{+}\}} \exp \left({\operatorname{sim}\left(\tilde A^j, \tilde A^n\right) / \tau_2 }\right)}\\
\end{aligned}
$$
where $\tilde A$ is the Gaussian smoothed version of the attention map, $\tau_1, \tau_2$ are temperature parameters, and $(j, j^+)$ and $(j, j^-)$ are positive and negative pairs.

The intuition behind using Gaussian smoothing is that it adds weights to the outside borders of attention regions, and the overlapping of outside borders is desired. The effect of the weights of attention regions in link loss, which is unwanted, is hopefully mostly counter-effected by contrast loss. In practice, we use Gaussian smoothing also for $A^j$
but with smaller kernel and variance compared to $\tilde{A}^j $

For contrast loss, given a token ($j$) in a noun group, we positively pair it with tokens in the same noun group  ($j^+$), and negatively pair it with tokens in different noun groups ($j^-$). Given an action token ($j$), we positively pair it with tokens of the interacting parties ($j^+$), and negatively pair it with tokens of other actions and other noun groups ($j^-$). We denote $\bar{\mathcal{L}}_\text{Contrast}$ as the average contrast loss of all positive pairs.

For link loss, we set the positive pairs as tokens in the same noun group and negative pairs as nouns in different noun groups but linked by a hyperedge (action). We denote $\bar{\mathcal{L}}_\text{Link}$ as the average link loss of all negative pairs.

The final loss function is then the weighted average of $\bar{\mathcal{L}}_\text{Contrast}$ and $\bar{\mathcal{L}}_\text{Link}$. The weights are controlled by iterative loss selection (0 or 1 on different
terms in the losses to perform iterative loss selection, c.f.\ Subsection \ref{subsec:shy})

\subsection{Conversation with ChatGPT-4-Turbo to obtain test data}
\label{subsec:conversation}

\textbf{User}

Generate seven prompts for image generation test of the rough format: object action object environment. The action should be involving two objects and evidently perceivable. The environment should be concise (preposition+noun). The words in the prompts should not be split by the tokenizer.

\textbf{ChatGPT}

Here are seven prompts adhering to your format:

\begin{enumerate}
    \item Robot painting canvas in studio
    \item Cat chasing mouse under tree
    \item Dog carrying stick over bridge
    \item Child reading book by fireplace
    \item Bird building nest on cliff
    \item Fisherman catching fish near river
    \item Artist sculpting statue in gallery
\end{enumerate}

\textbf{User}

Generate seven prompts for image generation test of the rough format: object action object environment. The objects may be described by adjectives. The action should be involving two objects and evidently perceivable. The environment should be concise (preposition+noun). The words in the prompts should not be split by the tokenizer.

\textbf{ChatGPT}

Here are seven prompts with adjectives describing the objects:

\begin{enumerate}
    \item Red robot fixing broken car in garage
    \item Small cat chasing gray mouse under table
    \item Golden retriever holding wooden stick on beach
    \item Young girl reading thick book by window
    \item Blue bird weaving tiny nest on branch
    \item Old fisherman reeling big fish near lake
    \item Skilled artist carving marble statue in workshop
\end{enumerate}

\section{Additional Figures}

\begin{figure}[H]
    \centering
    \includegraphics[width=0.5\linewidth]{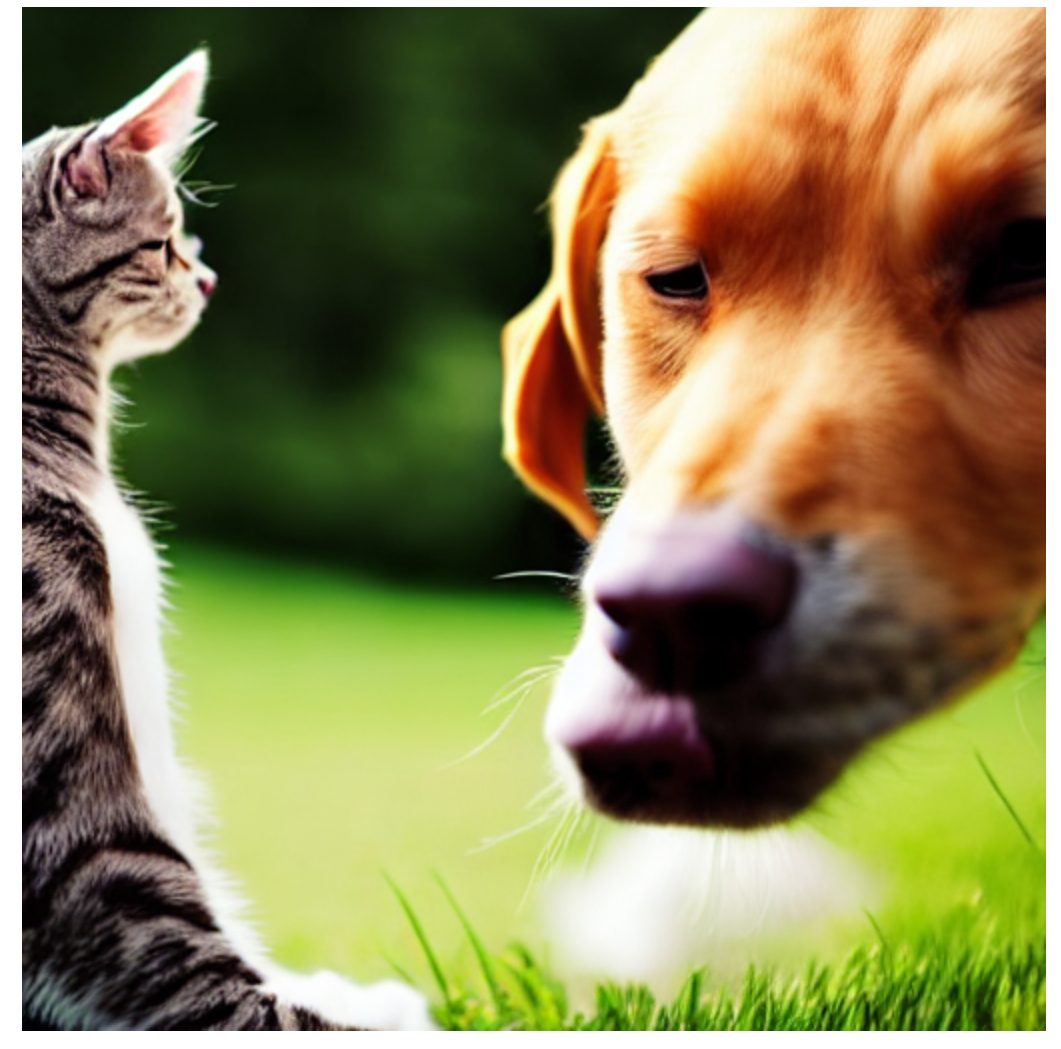}
    \caption{CONFORM with prompt: ``a cat playing with a dog" where (cat) contrasts with (dog); the action is not present}
    \label{fig:cat_dog}
\end{figure}

\begin{figure}[H]
    \centering
    \includegraphics[width=0.5\linewidth]{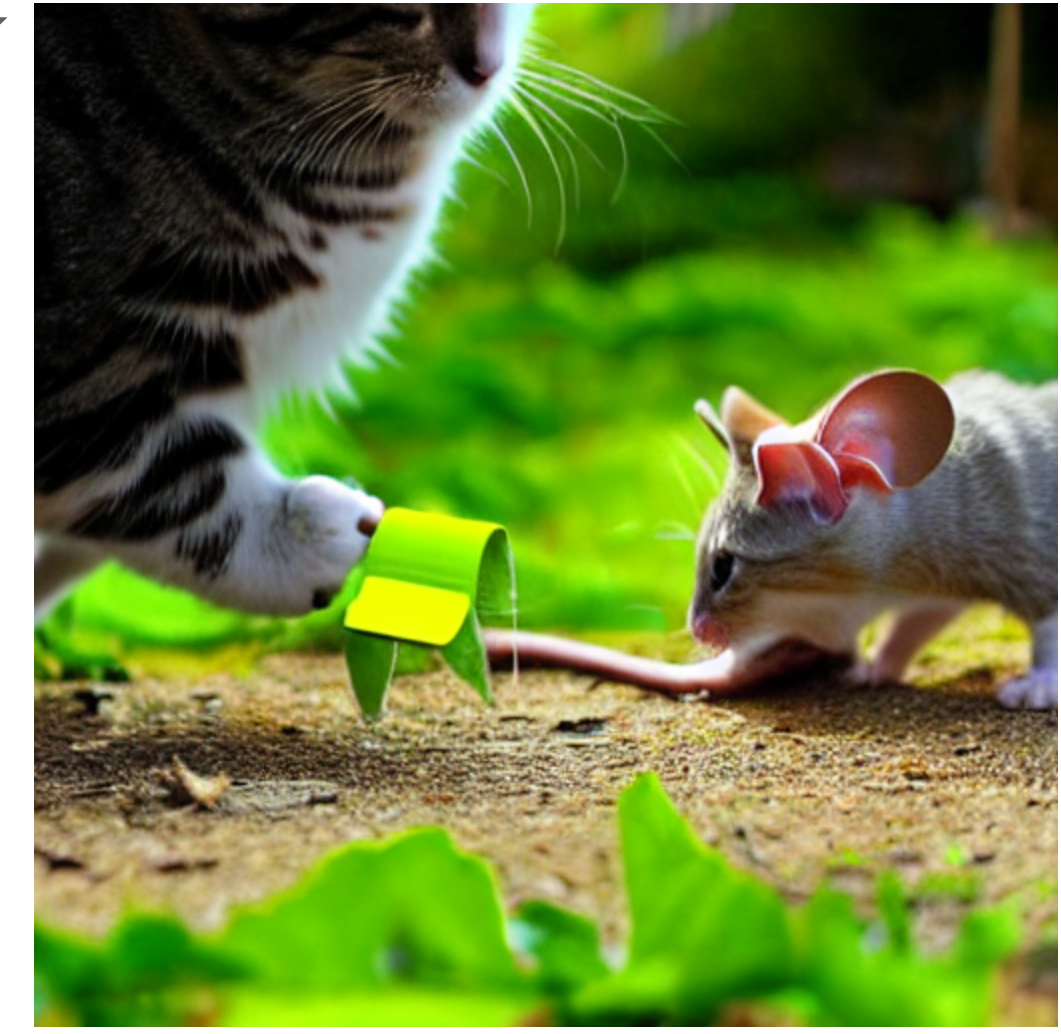}
    \caption{CONFORM with prompt: ``a cat catching a mouse" where (cat) contrasts with (mouse); the action is not present}
    \label{fig:cat_mouse}
\end{figure}

\begin{figure}[H]
    \centering
    \includegraphics[width=0.5\linewidth]{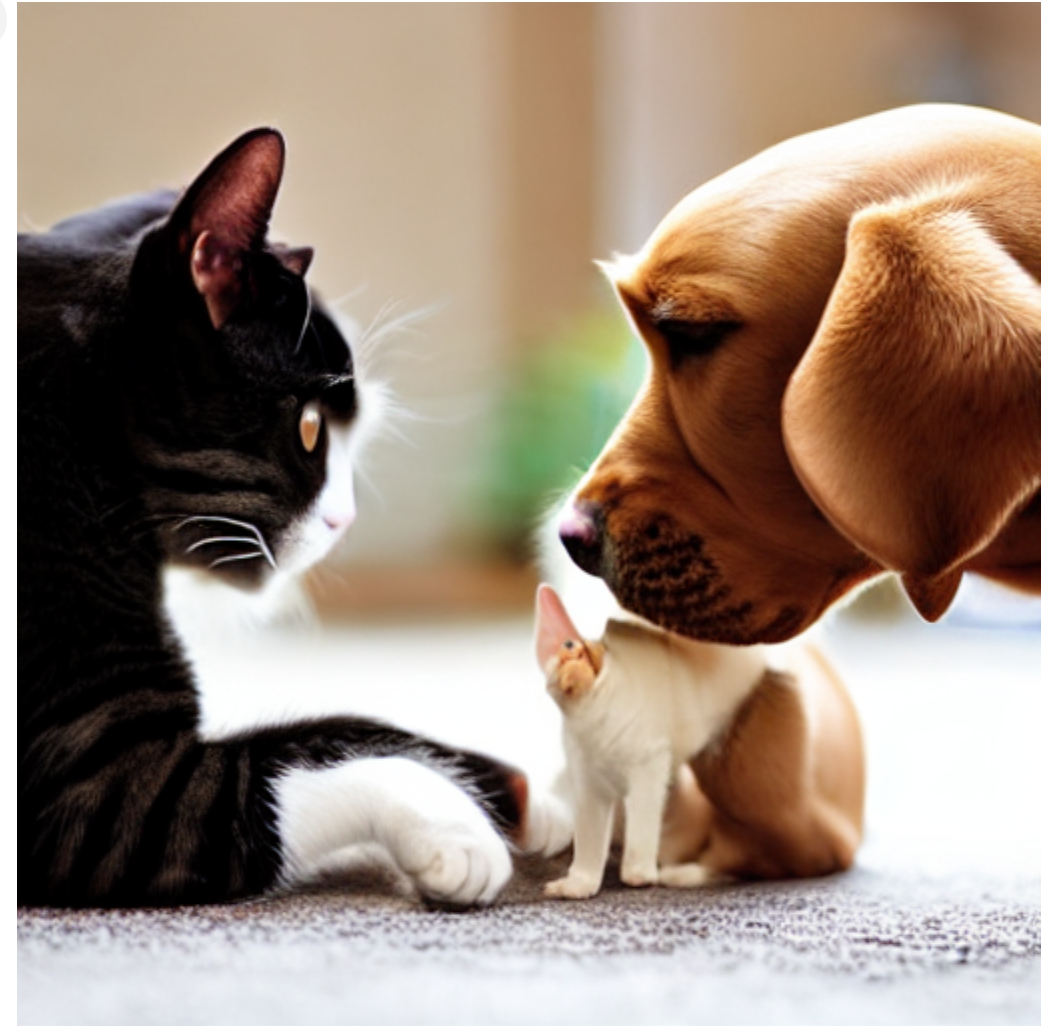}
    \caption{CONFORM with prompt: ``a cat playing with a dog" where (cat, playing) contrasts with (playing, dog); the objects are confused}
    \label{fig:cat_dog_obj}
\end{figure}

\begin{figure}[H]
    \centering
    \includegraphics[width=0.5\linewidth]{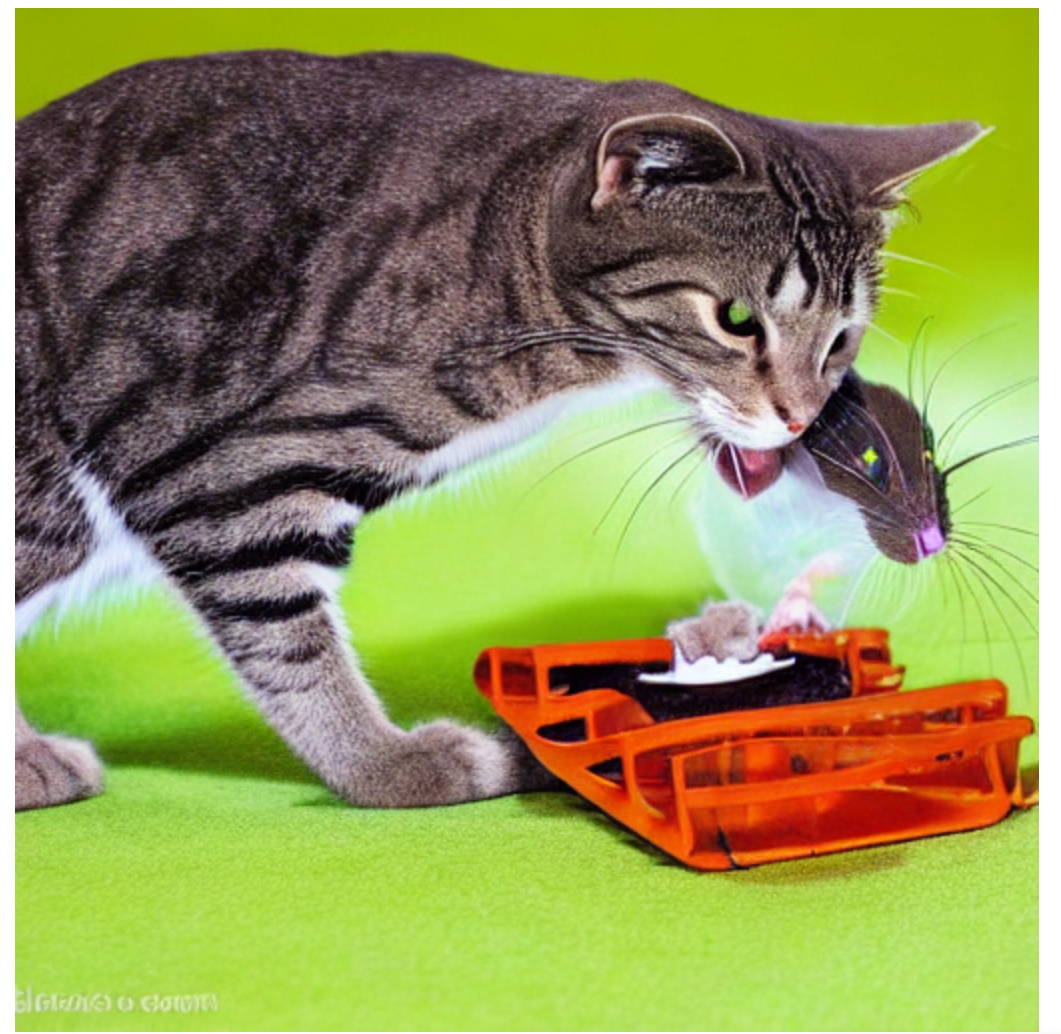}
    \caption{CONFORM with prompt: ``a cat catching a mouse" where (cat, catching) contrasts with (catching, mouse); the objects are confused}
    \label{fig:cat_mouse_obj}
\end{figure}

\begin{figure}
    \centering
    \includegraphics[width=\linewidth]{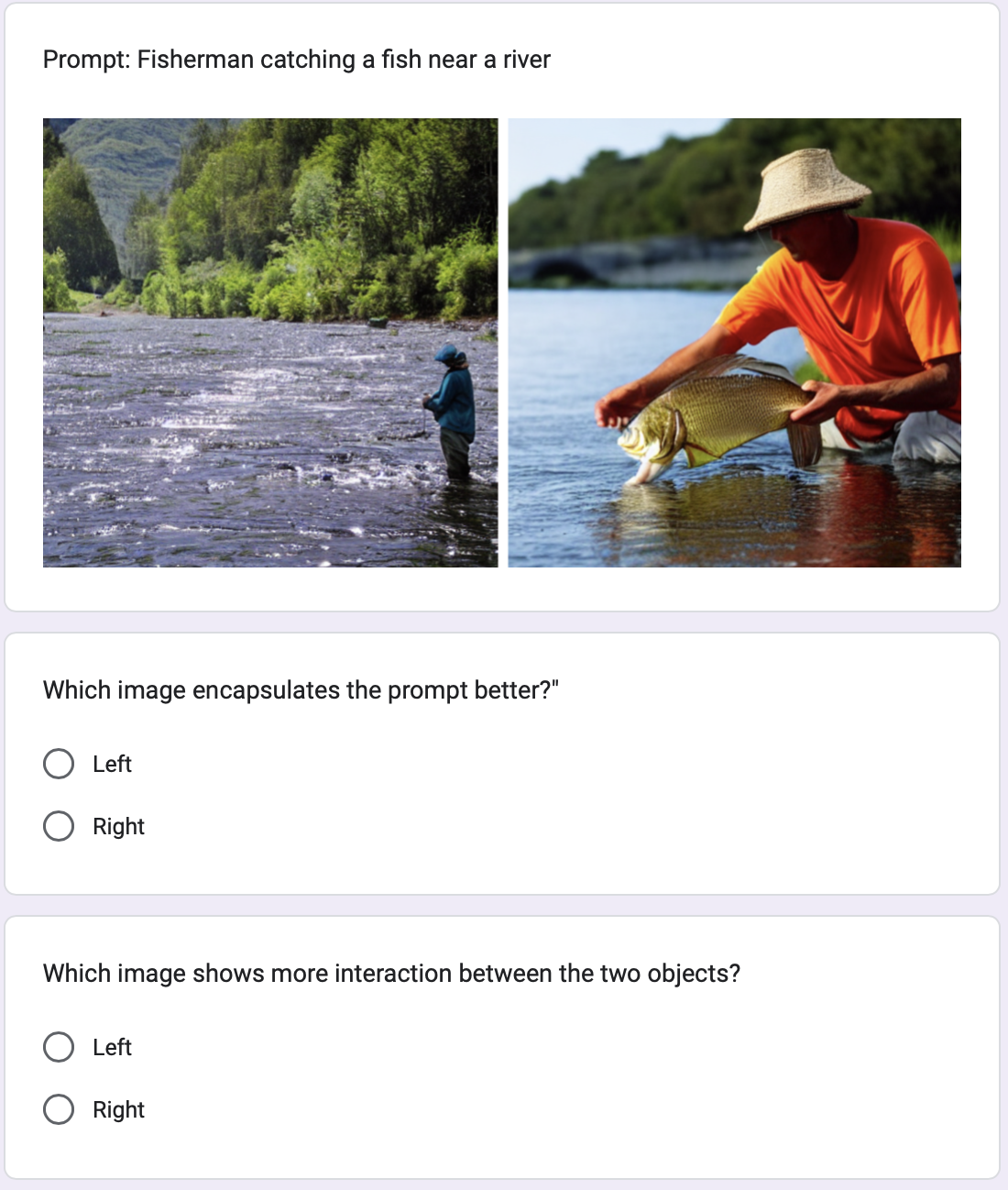}
    \caption{User study form example}
    \label{fig:google_form}
\end{figure}

\newpage

\subsection{Tests on complete implementations}
\label{subsec:collection}

Each image is labeled with user study scores (if available),
a TIFA score (question-answer pairs are generated by GPT-4o) and a CLIP score.

The user study scores comprise two values: the first is the percentage of people who voted for
``Which image encapsulates the prompt better?", the second
``Which image shows more interaction between the two objects?" of the same seed and prompt but for two models.

\begin{table}[h]
\centering
\begin{tabular}{|c|c|c|}
\hline
\textbf{Seed} & CONFORM  & SHYI  \\
\hline
4913 & \begin{tabular}{c} \tiny{USER: 6.9\%, 10.3\%} \\[-0.2cm] \tiny{TIFA: 1.0} \\[-0.2cm] \tiny{CLIP: 0.30} \\ \includegraphics[width=2.7cm]{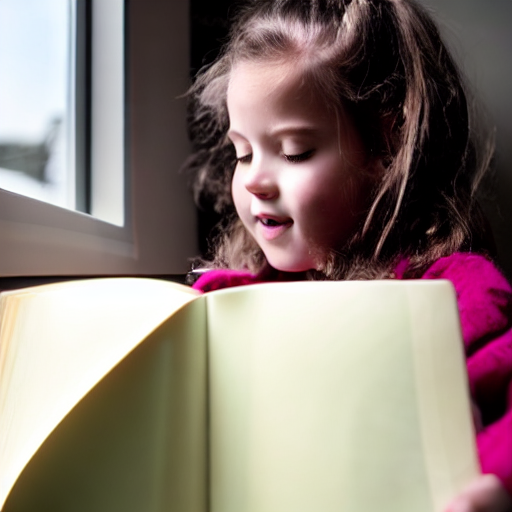} \end{tabular} & \begin{tabular}{c} \tiny{USER: 93.1\%, 89.7\%} \\[-0.2cm] \tiny{TIFA: 1.0} \\[-0.2cm] \tiny{CLIP: 0.31} \\ \includegraphics[width=2.7cm]{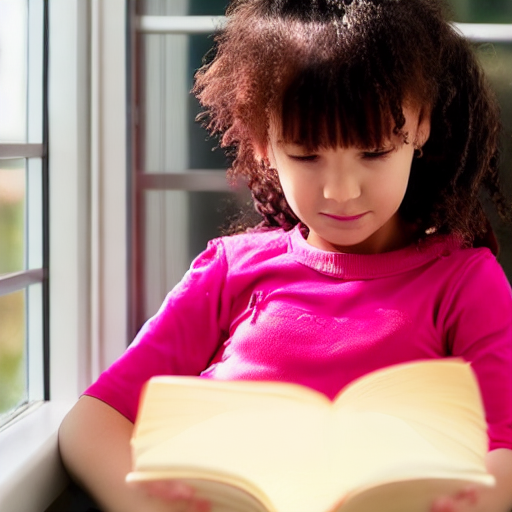} \end{tabular} \\
\hline
4914 & \begin{tabular}{c} \tiny{USER: 3.4\%, 10.3\%} \\[-0.2cm] \tiny{TIFA: 0.78} \\[-0.2cm] \tiny{CLIP: 0.29} \\ \includegraphics[width=2.7cm]{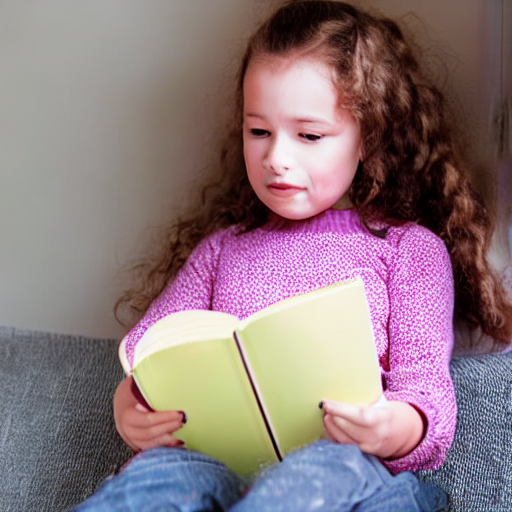} \end{tabular} & \begin{tabular}{c} \tiny{USER: 96.6\%, 89.7\%} \\[-0.2cm] \tiny{TIFA: 1.0} \\[-0.2cm] \tiny{CLIP: 0.32} \\ \includegraphics[width=2.7cm]{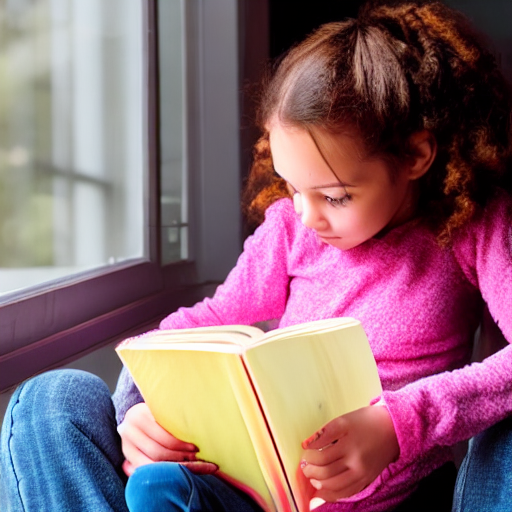} \end{tabular} \\
\hline
4915 & \begin{tabular}{c} \tiny{USER: 58.6\%, 34.5\%} \\[-0.2cm] \tiny{TIFA: 1.0} \\[-0.2cm] \tiny{CLIP: 0.30} \\ \includegraphics[width=2.7cm]{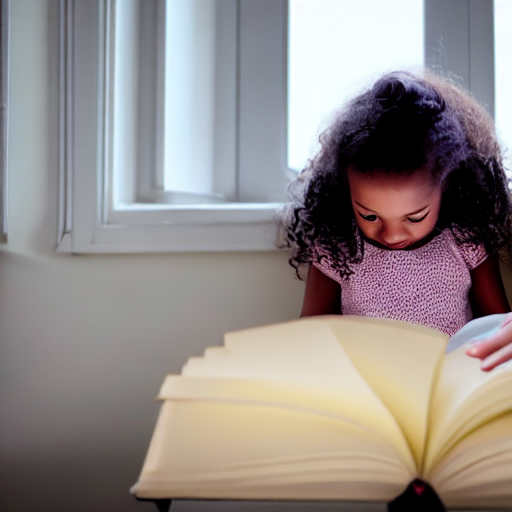} \end{tabular} & \begin{tabular}{c} \tiny{USER: 41.4\%, 65.5\%} \\[-0.2cm] \tiny{TIFA: 0.89} \\[-0.2cm] \tiny{CLIP: 0.31} \\ \includegraphics[width=2.7cm]{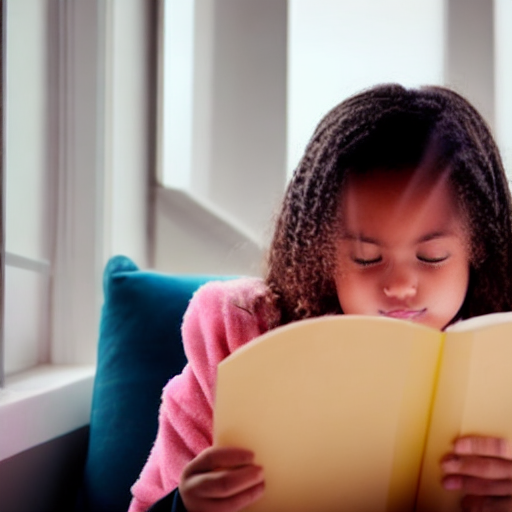} \end{tabular} \\
\hline
\end{tabular}
\caption{Prompt: \textit{Young girl reading thick book by window}}
\label{tab:Young girl reading thick book by window}
\end{table}

\begin{table}[h]
\centering
\begin{tabular}{|c|c|c|}
\hline
\textbf{Seed} & CONFORM  & SHYI  \\
\hline
4913 & \begin{tabular}{c} \tiny{USER: 67.9\%, 53.6\%} \\[-0.2cm] \tiny{TIFA: 1.0} \\[-0.2cm] \tiny{CLIP: 0.36} \\ \includegraphics[width=2.7cm]{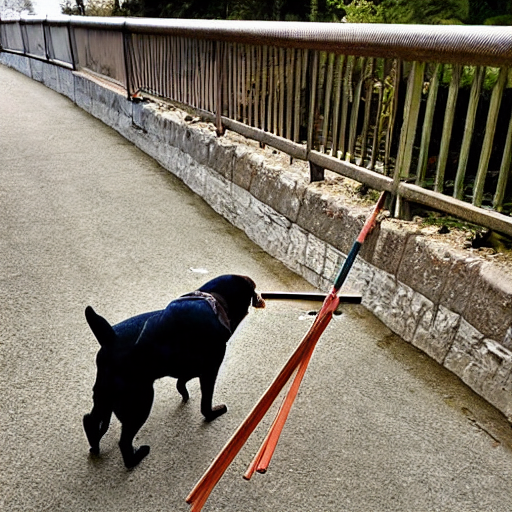} \end{tabular} & \begin{tabular}{c} \tiny{USER: 32.1\%, 46.4\%} \\[-0.2cm] \tiny{TIFA: 1.0} \\[-0.2cm] \tiny{CLIP: 0.33} \\ \includegraphics[width=2.7cm]{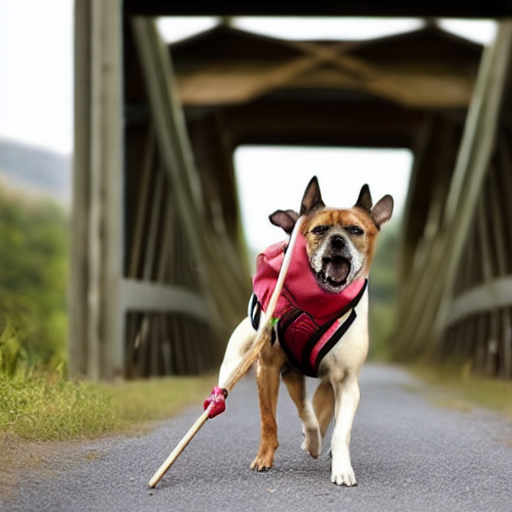} \end{tabular} \\
\hline
4914 & \begin{tabular}{c} \tiny{USER: 67.9\%, 78.6\%} \\[-0.2cm] \tiny{TIFA: 1.0} \\[-0.2cm] \tiny{CLIP: 0.33} \\ \includegraphics[width=2.7cm]{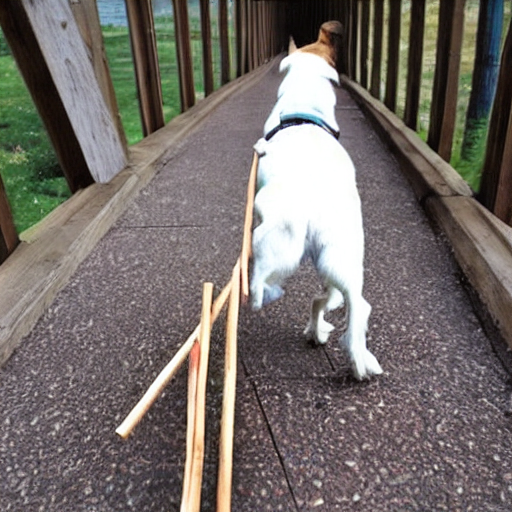} \end{tabular} & \begin{tabular}{c} \tiny{USER: 32.1\%, 21.4\%} \\[-0.2cm] \tiny{TIFA: 0.75} \\[-0.2cm] \tiny{CLIP: 0.33} \\ \includegraphics[width=2.7cm]{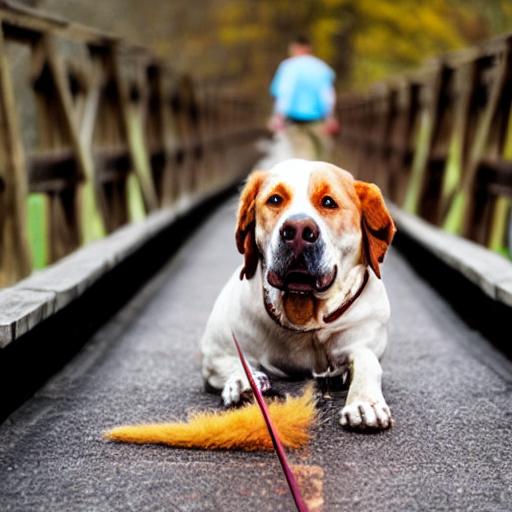} \end{tabular} \\
\hline
4915 & \begin{tabular}{c} \tiny{USER: 0\%, 11.5\%} \\[-0.2cm] \tiny{TIFA: 0.50} \\[-0.2cm] \tiny{CLIP: 0.27} \\ \includegraphics[width=2.7cm]{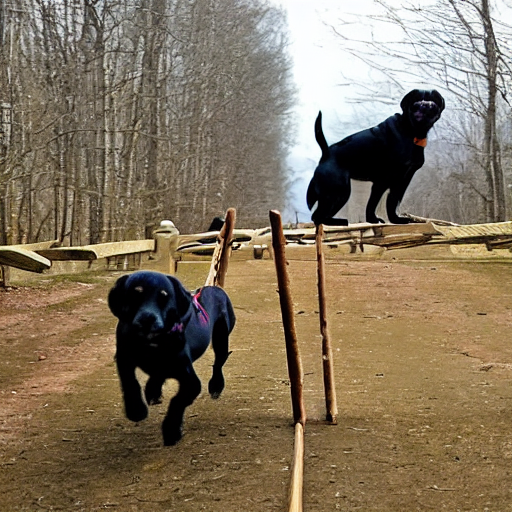} \end{tabular} & \begin{tabular}{c} \tiny{USER: 100\%, 88.5\%} \\[-0.2cm] \tiny{TIFA: 1.0} \\[-0.2cm] \tiny{CLIP: 0.32} \\ \includegraphics[width=2.7cm]{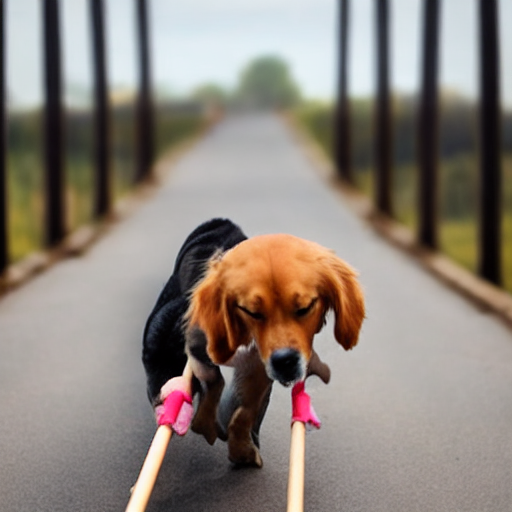} \end{tabular} \\
\hline
\end{tabular}
\caption{Prompt: \textit{Dog carrying stick over bridge}}
\label{tab:Dog carrying stick over bridge}
\end{table}

\begin{table}[h]
\centering
\begin{tabular}{|c|c|c|}
\hline
\textbf{Seed} & CONFORM  & SHYI  \\
\hline
4913 & \begin{tabular}{c} \tiny{USER: 14.3\%, 14.8\%} \\[-0.2cm] \tiny{TIFA: 1.0} \\[-0.2cm] \tiny{CLIP: 0.26} \\ \includegraphics[width=2.7cm]{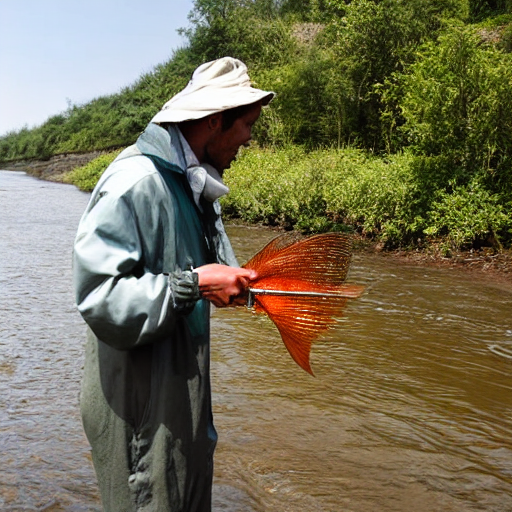} \end{tabular} & \begin{tabular}{c} \tiny{USER: 85.7\%, 85.2\%} \\[-0.2cm] \tiny{TIFA: 1.0} \\[-0.2cm] \tiny{CLIP: 0.26} \\ \includegraphics[width=2.7cm]{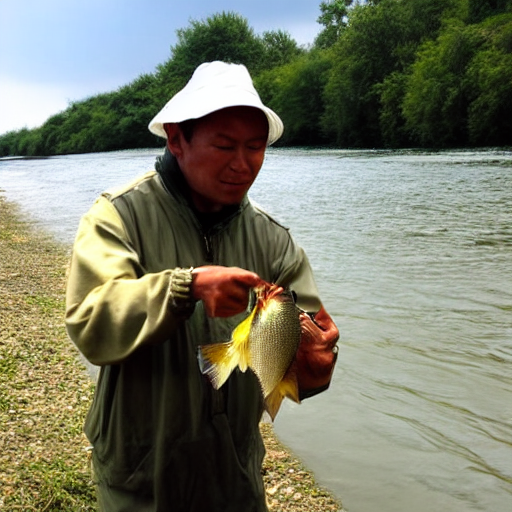} \end{tabular} \\
\hline
4914 & \begin{tabular}{c} \tiny{USER: 28.6\%, 14.3\%} \\[-0.2cm] \tiny{TIFA: 1.0} \\[-0.2cm] \tiny{CLIP: 0.28} \\ \includegraphics[width=2.7cm]{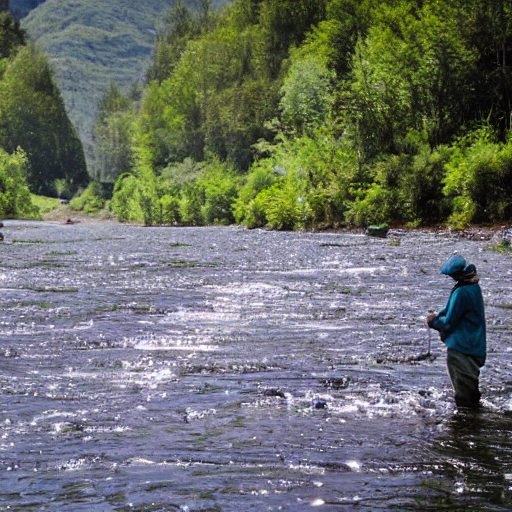} \end{tabular} & \begin{tabular}{c} \tiny{USER: 71.4\%, 85.7\%} \\[-0.2cm] \tiny{TIFA: 0.83} \\[-0.2cm] \tiny{CLIP: 0.28} \\ \includegraphics[width=2.7cm]{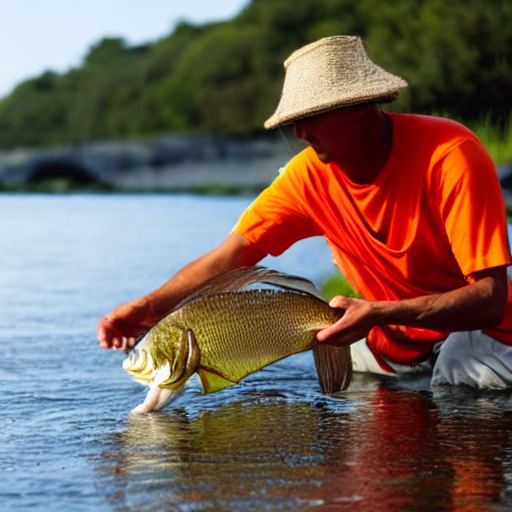} \end{tabular} \\
\hline
4915 & \begin{tabular}{c} \tiny{USER: 32.1\%, 32.1\%} \\[-0.2cm] \tiny{TIFA: 0.83} \\[-0.2cm] \tiny{CLIP: 0.28} \\ \includegraphics[width=2.7cm]{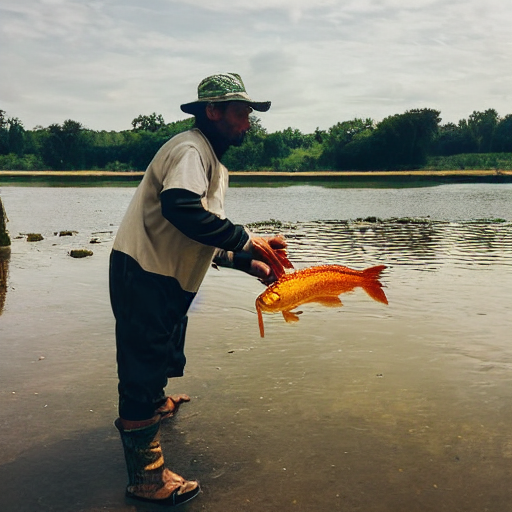} \end{tabular} & \begin{tabular}{c} \tiny{USER: 67.9\%, 67.9\%} \\[-0.2cm] \tiny{TIFA: 0.83} \\[-0.2cm] \tiny{CLIP: 0.27} \\ \includegraphics[width=2.7cm]{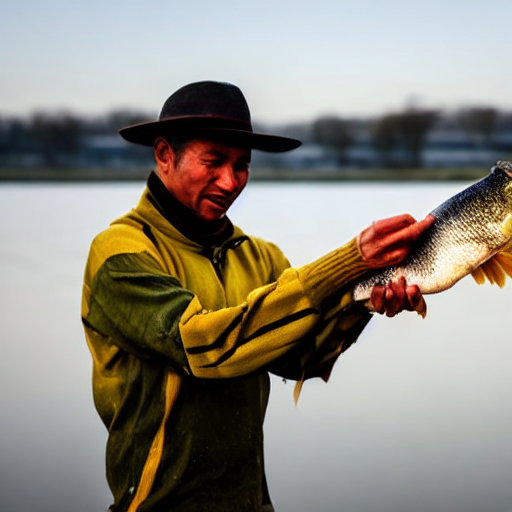} \end{tabular} \\
\hline
\end{tabular}
\caption{Prompt: \textit{Fisherman catching fish near river}}
\label{tab:Fisherman catching fish near river}
\end{table}

\begin{table}[h]
\centering
\begin{tabular}{|c|c|c|}
\hline
\textbf{Seed} & CONFORM  & SHYI  \\
\hline
4913 & \begin{tabular}{c} \tiny{USER: 51.9\%, 71.4\%} \\[-0.2cm] \tiny{TIFA: 0.87} \\[-0.2cm] \tiny{CLIP: 0.35} \\ \includegraphics[width=2.7cm]{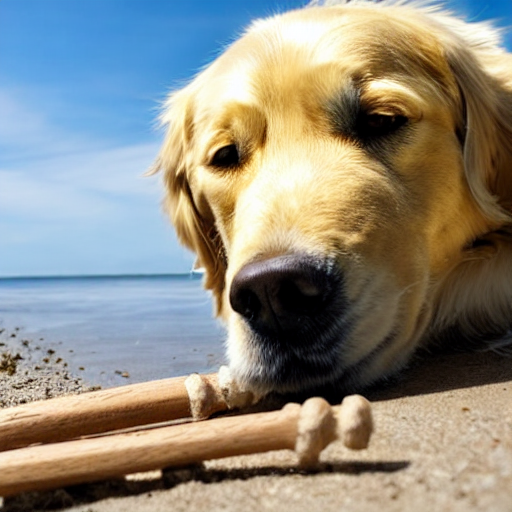} \end{tabular} & \begin{tabular}{c} \tiny{USER: 48.1\%, 28.6\%} \\[-0.2cm] \tiny{TIFA: 0.87} \\[-0.2cm] \tiny{CLIP: 0.37} \\ \includegraphics[width=2.7cm]{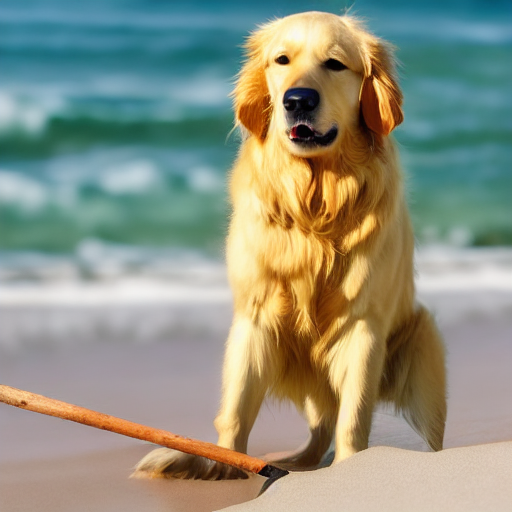} \end{tabular} \\
\hline
4914 & \begin{tabular}{c} \tiny{USER: 32.1\%, 48.1\%} \\[-0.2cm] \tiny{TIFA: 0.87} \\[-0.2cm] \tiny{CLIP: 0.35} \\ \includegraphics[width=2.7cm]{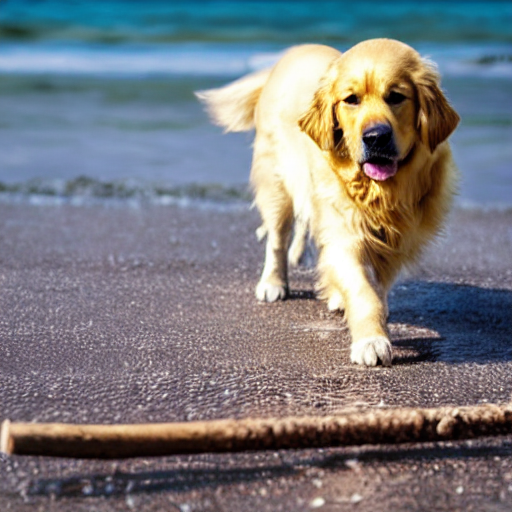} \end{tabular} & \begin{tabular}{c} \tiny{USER: 67.9\%, 51.9\%} \\[-0.2cm] \tiny{TIFA: 0.87} \\[-0.2cm] \tiny{CLIP: 0.33} \\ \includegraphics[width=2.7cm]{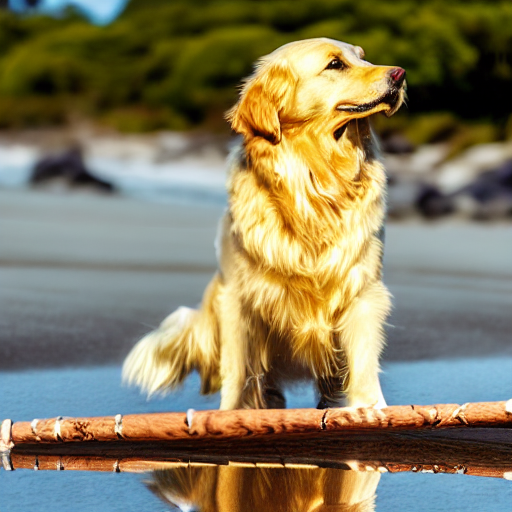} \end{tabular} \\
\hline
4915 & \begin{tabular}{c} \tiny{USER: 85.2\%, 61.5\%} \\[-0.2cm] \tiny{TIFA: 1.0} \\[-0.2cm] \tiny{CLIP: 0.39} \\ \includegraphics[width=2.7cm]{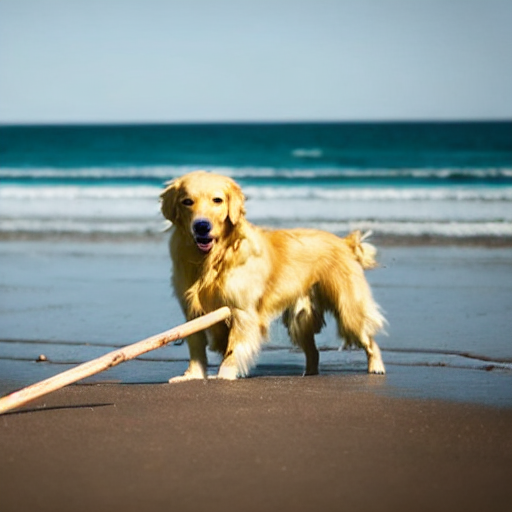} \end{tabular} & \begin{tabular}{c} \tiny{USER: 14.8\%, 38.5\%} \\[-0.2cm] \tiny{TIFA: 1.0} \\[-0.2cm] \tiny{CLIP: 0.38} \\ \includegraphics[width=2.7cm]{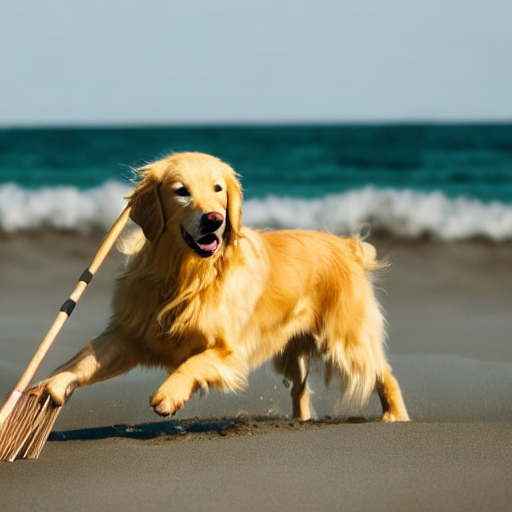} \end{tabular} \\
\hline
\end{tabular}
\caption{Prompt: \textit{Golden retriever holding wooden stick on beach}}
\label{tab:Golden retriever holding wooden stick on beach}
\end{table}

\begin{table}[h]
\centering
\begin{tabular}{|c|c|c|}
\hline
\textbf{Seed} & CONFORM  & SHYI  \\
\hline
4913 & \begin{tabular}{c} \tiny{USER: 60.7\%, 64.3\%} \\[-0.2cm] \tiny{TIFA: 1.0} \\[-0.2cm] \tiny{CLIP: 0.33} \\ \includegraphics[width=2.7cm]{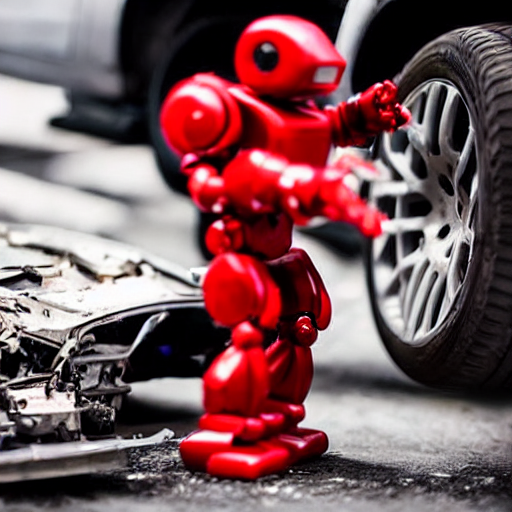} \end{tabular} & \begin{tabular}{c} \tiny{USER: 39.3\%, 35.7\%} \\[-0.2cm] \tiny{TIFA: 0.82} \\[-0.2cm] \tiny{CLIP: 0.32} \\ \includegraphics[width=2.7cm]{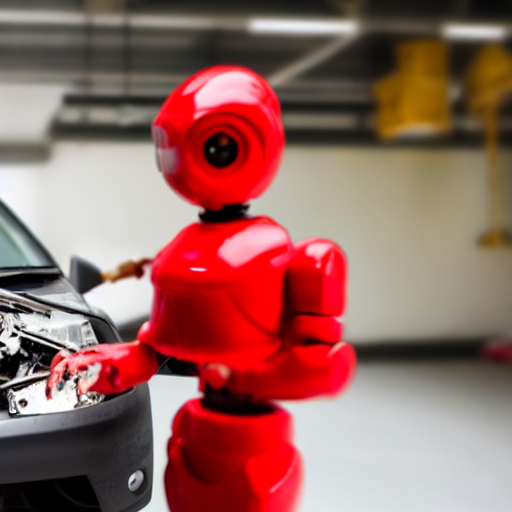} \end{tabular} \\
\hline
4914 & \begin{tabular}{c} \tiny{USER: 89.3\%, 88.9\%} \\[-0.2cm] \tiny{TIFA: 0.91} \\[-0.2cm] \tiny{CLIP: 0.28} \\ \includegraphics[width=2.7cm]{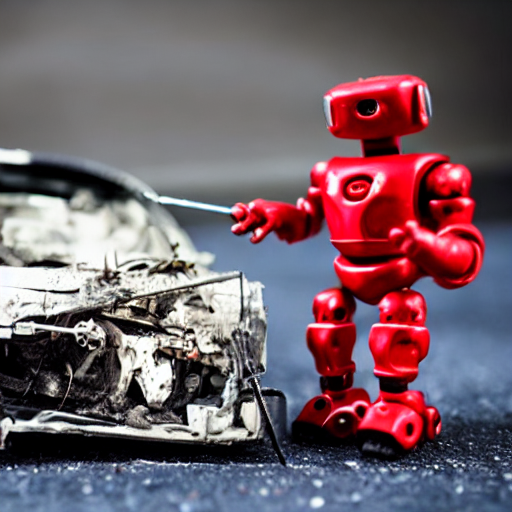} \end{tabular} & \begin{tabular}{c} \tiny{USER: 10.7\%, 11.1\%} \\[-0.2cm] \tiny{TIFA: 0.91} \\[-0.2cm] \tiny{CLIP: 0.28} \\ \includegraphics[width=2.7cm]{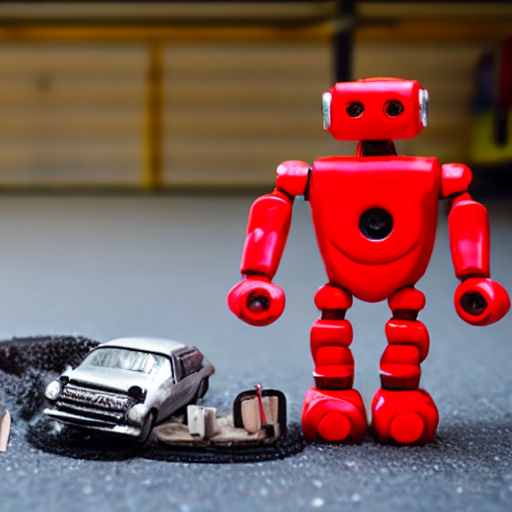} \end{tabular} \\
\hline
4915 & \begin{tabular}{c} \tiny{USER: 53.6\%, 28.6\%} \\[-0.2cm] \tiny{TIFA: 0.73} \\[-0.2cm] \tiny{CLIP: 0.29} \\ \includegraphics[width=2.7cm]{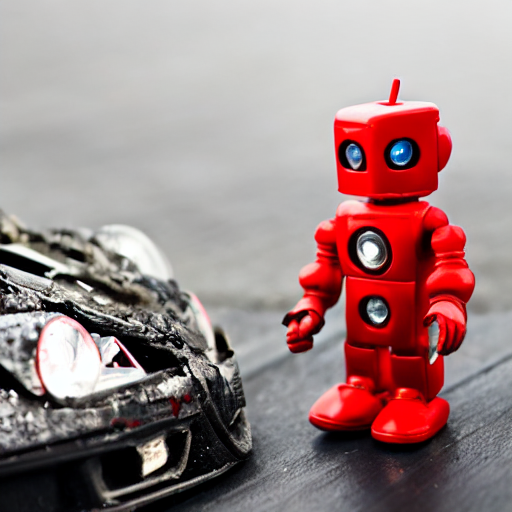} \end{tabular} & \begin{tabular}{c} \tiny{USER: 46.4\%, 71.4\%} \\[-0.2cm] \tiny{TIFA: 0.82} \\[-0.2cm] \tiny{CLIP: 0.29} \\ \includegraphics[width=2.7cm]{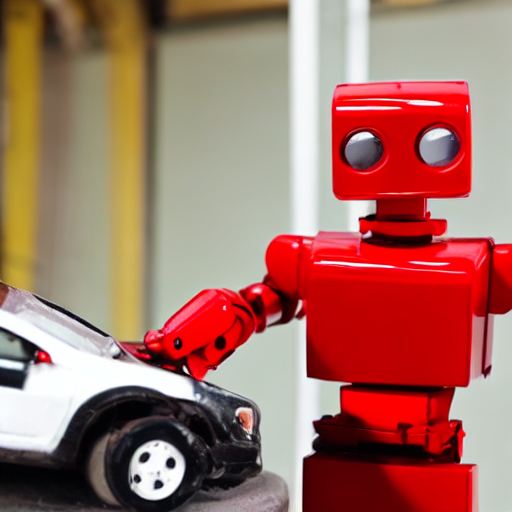} \end{tabular} \\
\hline
\end{tabular}
\caption{Prompt: \textit{Red robot fixing broken car in garage}}
\label{tab:Red robot fixing broken car in garage}
\end{table}

\begin{table}[h]
\centering
\begin{tabular}{|c|c|c|}
\hline
\textbf{Seed} & CONFORM  & SHYI  \\
\hline
4913 & \begin{tabular}{c} \tiny{USER: 57.1\%, 78.6\%} \\[-0.2cm] \tiny{TIFA: 1.0} \\[-0.2cm] \tiny{CLIP: 0.33} \\ \includegraphics[width=2.7cm]{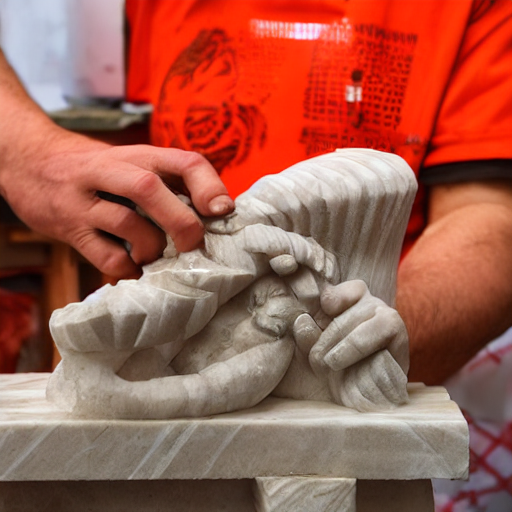} \end{tabular} & \begin{tabular}{c} \tiny{USER: 42.9\%, 21.4\%} \\[-0.2cm] \tiny{TIFA: 0.83} \\[-0.2cm] \tiny{CLIP: 0.34} \\ \includegraphics[width=2.7cm]{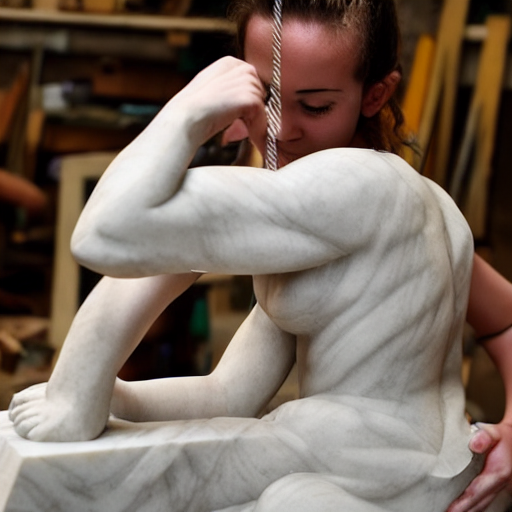} \end{tabular} \\
\hline
4914 & \begin{tabular}{c} \tiny{USER: 48.7\%, 57.1\%} \\[-0.2cm] \tiny{TIFA: 1.0} \\[-0.2cm] \tiny{CLIP: 0.35} \\ \includegraphics[width=2.7cm]{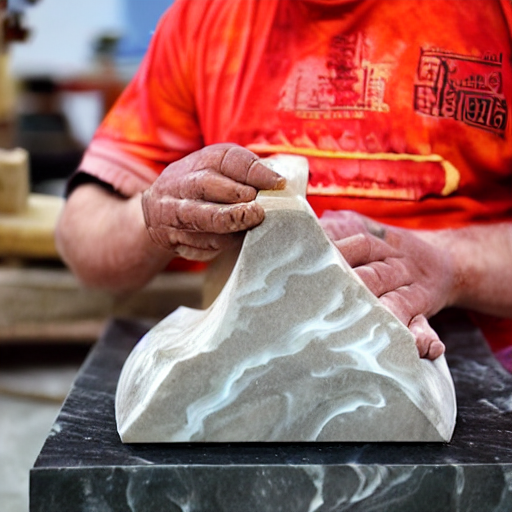} \end{tabular} & \begin{tabular}{c} \tiny{USER: 51.3\%, 42.9\%} \\[-0.2cm] \tiny{TIFA: 0.83} \\[-0.2cm] \tiny{CLIP: 0.31} \\ \includegraphics[width=2.7cm]{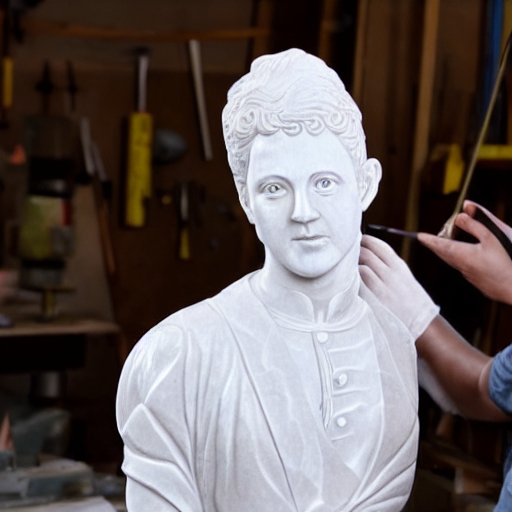} \end{tabular} \\
\hline
4915 & \begin{tabular}{c} \tiny{USER: 66.7\%, 63\%} \\[-0.2cm] \tiny{TIFA: 1.0} \\[-0.2cm] \tiny{CLIP: 0.34} \\ \includegraphics[width=2.7cm]{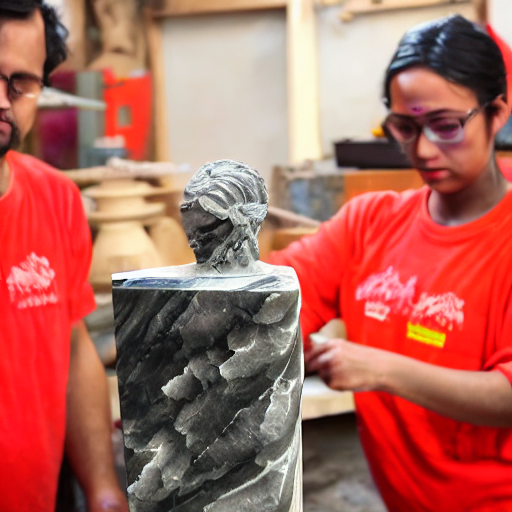} \end{tabular} & \begin{tabular}{c} \tiny{USER: 33.3\%, 37\%} \\[-0.2cm] \tiny{TIFA: 0.83} \\[-0.2cm] \tiny{CLIP: 0.34} \\ \includegraphics[width=2.7cm]{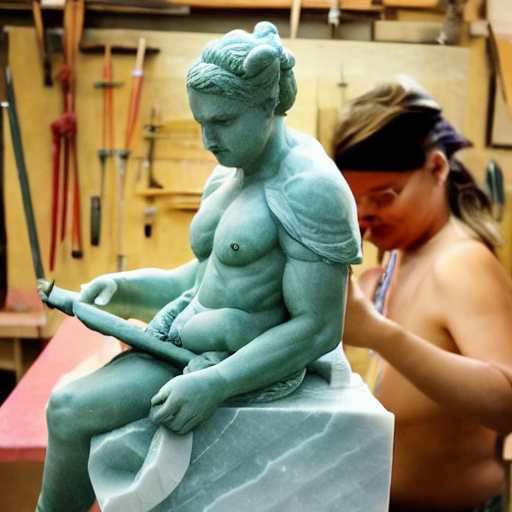} \end{tabular} \\
\hline
\end{tabular}
\caption{Prompt: \textit{Skilled artist carving marble statue in workshop}}
\label{tab:Skilled artist carving marble statue in workshop}
\end{table}

\begin{table}[h]
\centering
\begin{tabular}{|c|c|c|}
\hline
\textbf{Seed} & CONFORM  & SHYI  \\
\hline
4913 & \begin{tabular}{c} \tiny{USER: xx\%, xx\%} \\[-0.2cm] \tiny{TIFA: 0.45} \\[-0.2cm] \tiny{CLIP: 0.26} \\ \includegraphics[width=2.7cm]{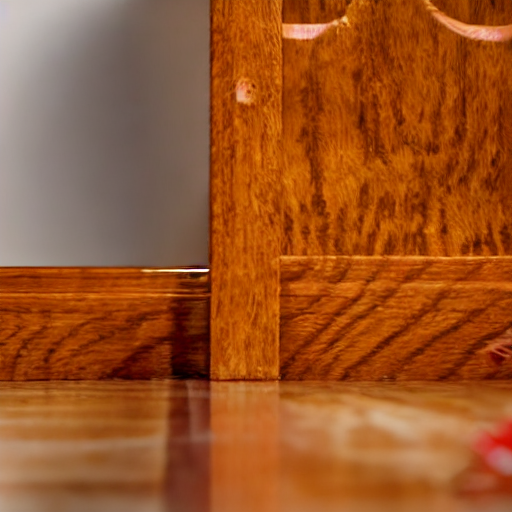} \end{tabular} & \begin{tabular}{c} \tiny{USER: xx\%, xx\%} \\[-0.2cm] \tiny{TIFA: 0.60} \\[-0.2cm] \tiny{CLIP: 0.29} \\ \includegraphics[width=2.7cm]{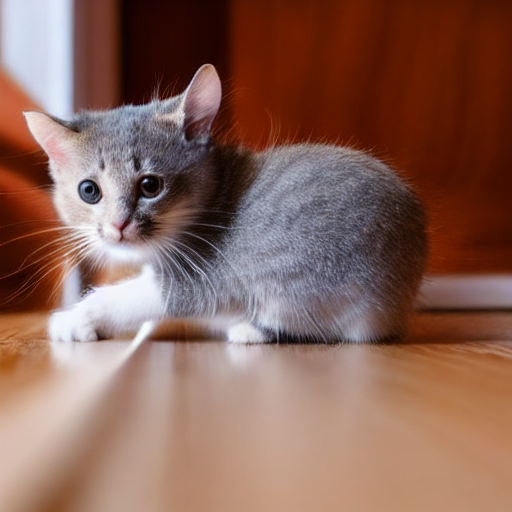} \end{tabular} \\
\hline
4914 & \begin{tabular}{c} \tiny{USER: xx\%, xx\%} \\[-0.2cm] \tiny{TIFA: 0.82} \\[-0.2cm] \tiny{CLIP: 0.31} \\ \includegraphics[width=2.7cm]{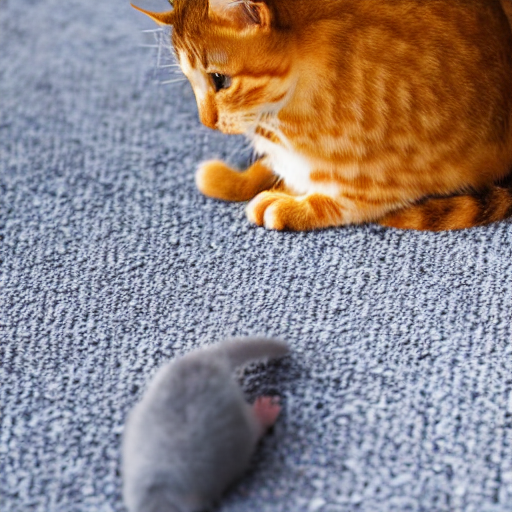} \end{tabular} & \begin{tabular}{c} \tiny{USER: xx\%, xx\%} \\[-0.2cm] \tiny{TIFA: 0.60} \\[-0.2cm] \tiny{CLIP: 0.25} \\ \includegraphics[width=2.7cm]{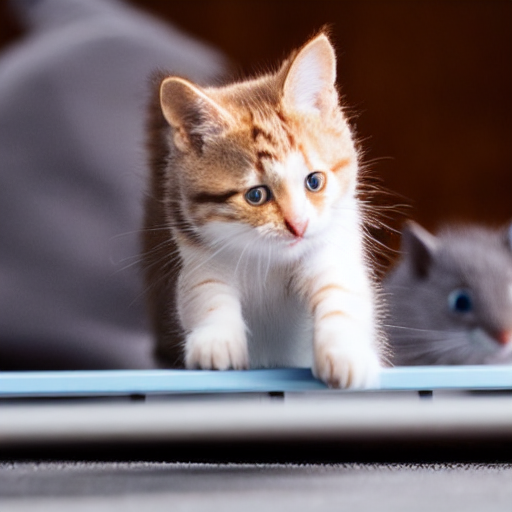} \end{tabular} \\
\hline
4915 & \begin{tabular}{c} \tiny{USER: xx\%, xx\%} \\[-0.2cm] \tiny{TIFA: 0.55} \\[-0.2cm] \tiny{CLIP: 0.29} \\ \includegraphics[width=2.7cm]{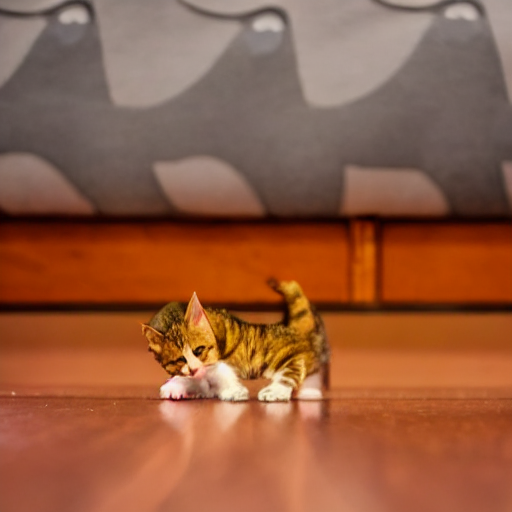} \end{tabular} & \begin{tabular}{c} \tiny{USER: xx\%, xx\%} \\[-0.2cm] \tiny{TIFA: 0.90} \\[-0.2cm] \tiny{CLIP: 0.32} \\ \includegraphics[width=2.7cm]{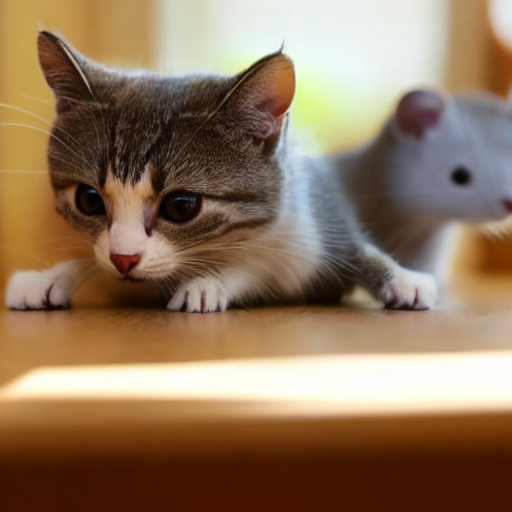} \end{tabular} \\
\hline
\end{tabular}
\caption{Prompt: \textit{Small cat chasing gray mouse under table}}
\label{tab:Small cat chasing gray mouse under table}
\end{table}

\begin{table}[h]
\centering
\begin{tabular}{|c|c|c|}
\hline
\textbf{Seed} & CONFORM  & SHYI  \\
\hline
4913 & \begin{tabular}{c} \tiny{USER: xx\%, xx\%} \\[-0.2cm] \tiny{TIFA: 1.0} \\[-0.2cm] \tiny{CLIP: 0.32} \\ \includegraphics[width=2.7cm]{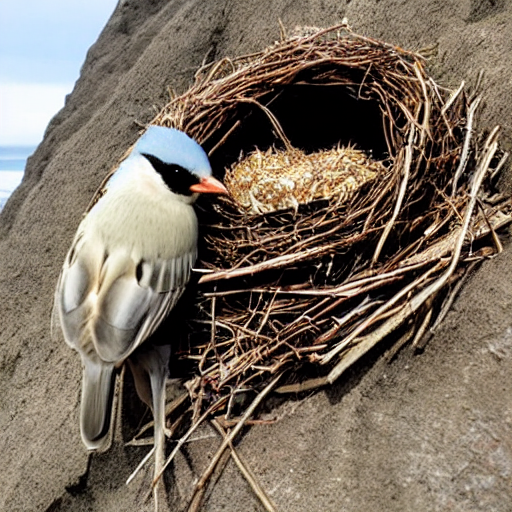} \end{tabular} & \begin{tabular}{c} \tiny{USER: xx\%, xx\%} \\[-0.2cm] \tiny{TIFA: 1.0} \\[-0.2cm] \tiny{CLIP: 0.34} \\ \includegraphics[width=2.7cm]{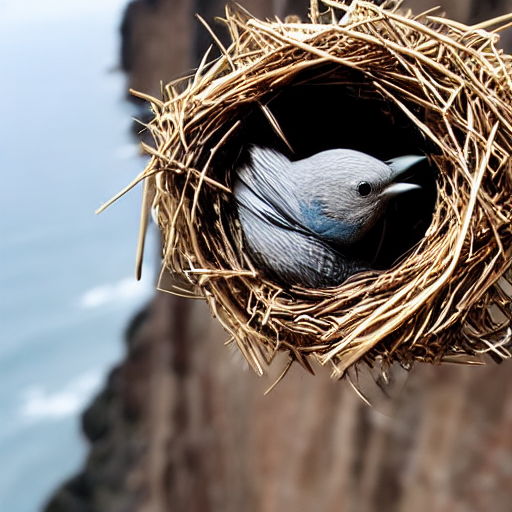} \end{tabular} \\
\hline
4914 & \begin{tabular}{c} \tiny{USER: xx\%, xx\%} \\[-0.2cm] \tiny{TIFA: 1.0} \\[-0.2cm] \tiny{CLIP: 0.30} \\ \includegraphics[width=2.7cm]{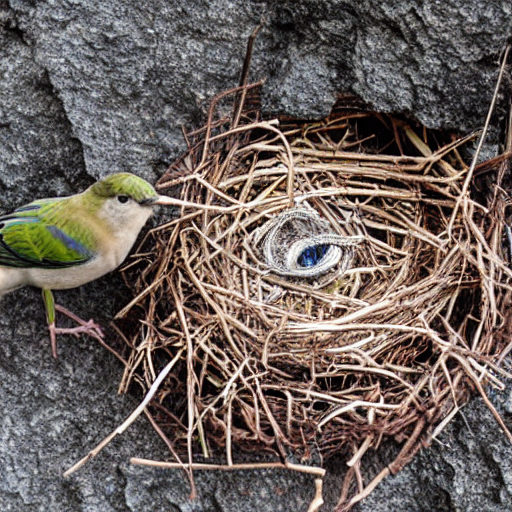} \end{tabular} & \begin{tabular}{c} \tiny{USER: xx\%, xx\%} \\[-0.2cm] \tiny{TIFA: 1.0} \\[-0.2cm] \tiny{CLIP: 0.33} \\ \includegraphics[width=2.7cm]{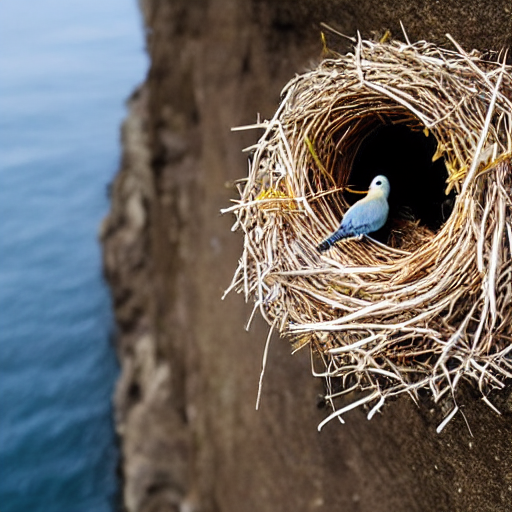} \end{tabular} \\
\hline
4915 & \begin{tabular}{c} \tiny{USER: xx\%, xx\%} \\[-0.2cm] \tiny{TIFA: 1.0} \\[-0.2cm] \tiny{CLIP: 0.31} \\ \includegraphics[width=2.7cm]{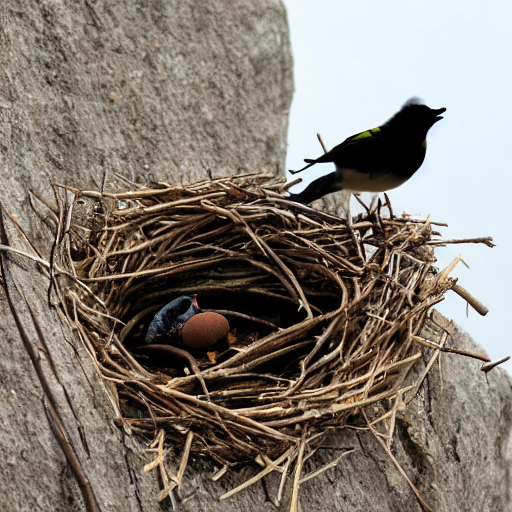} \end{tabular} & \begin{tabular}{c} \tiny{USER: xx\%, xx\%} \\[-0.2cm] \tiny{TIFA: 0.83} \\[-0.2cm] \tiny{CLIP: 0.28} \\ \includegraphics[width=2.7cm]{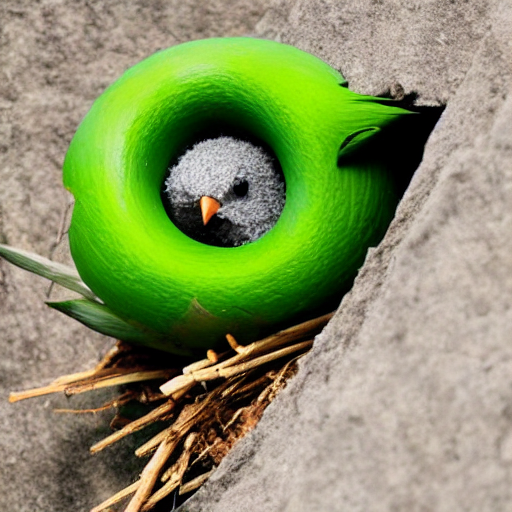} \end{tabular} \\
\hline
\end{tabular}
\caption{Prompt: \textit{Bird building nest on cliff}}
\label{tab:Bird building nest on cliff}
\end{table}

\begin{table}[h]
\centering
\begin{tabular}{|c|c|c|}
\hline
\textbf{Seed} & CONFORM  & SHYI  \\
\hline
4913 & \begin{tabular}{c} \tiny{USER: xx\%, xx\%} \\[-0.2cm] \tiny{TIFA: 1.0} \\[-0.2cm] \tiny{CLIP: 0.32} \\ \includegraphics[width=2.7cm]{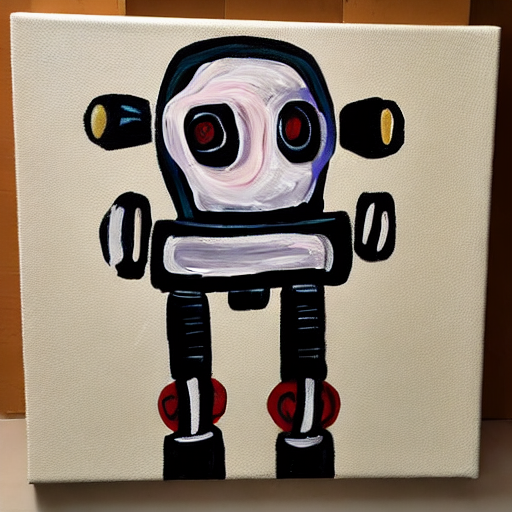} \end{tabular} & \begin{tabular}{c} \tiny{USER: xx\%, xx\%} \\[-0.2cm] \tiny{TIFA: 0.83} \\[-0.2cm] \tiny{CLIP: 0.29} \\ \includegraphics[width=2.7cm]{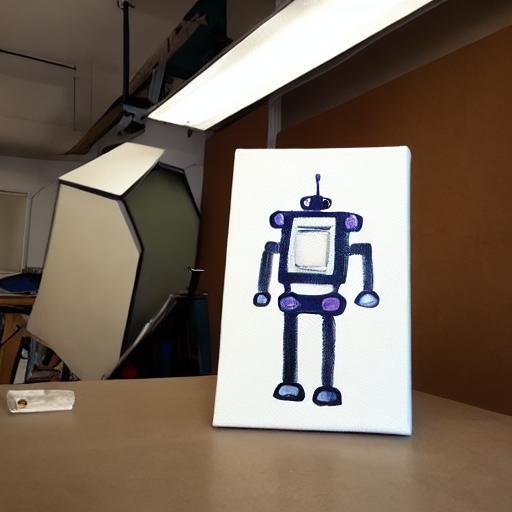} \end{tabular} \\
\hline
4914 & \begin{tabular}{c} \tiny{USER: xx\%, xx\%} \\[-0.2cm] \tiny{TIFA: 0.83} \\[-0.2cm] \tiny{CLIP: 0.33} \\ \includegraphics[width=2.7cm]{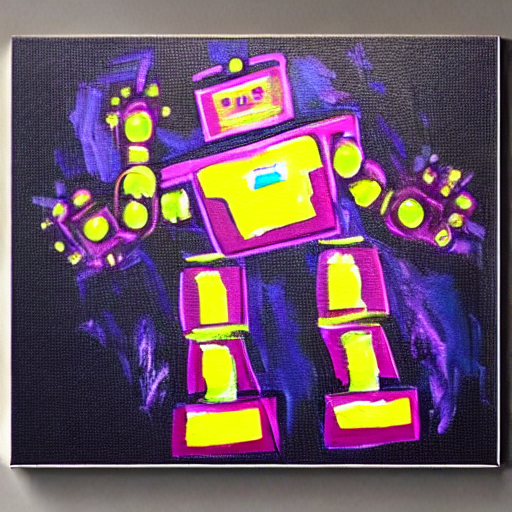} \end{tabular} & \begin{tabular}{c} \tiny{USER: xx\%, xx\%} \\[-0.2cm] \tiny{TIFA: 1.0} \\[-0.2cm] \tiny{CLIP: 0.33} \\ \includegraphics[width=2.7cm]{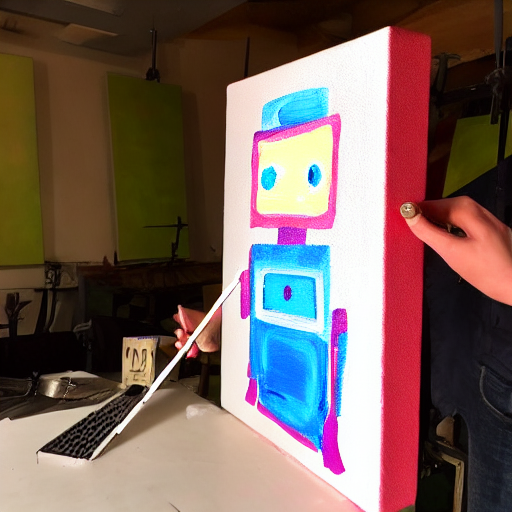} \end{tabular} \\
\hline
4915 & \begin{tabular}{c} \tiny{USER: xx\%, xx\%} \\[-0.2cm] \tiny{TIFA: 1.0} \\[-0.2cm] \tiny{CLIP: 0.34} \\ \includegraphics[width=2.7cm]{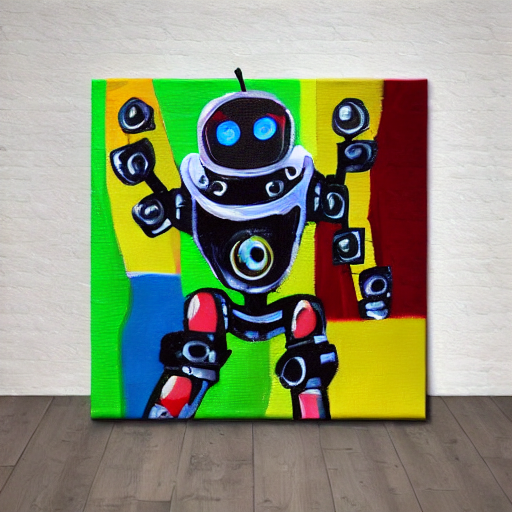} \end{tabular} & \begin{tabular}{c} \tiny{USER: xx\%, xx\%} \\[-0.2cm] \tiny{TIFA: 0.67} \\[-0.2cm] \tiny{CLIP: 0.31} \\ \includegraphics[width=2.7cm]{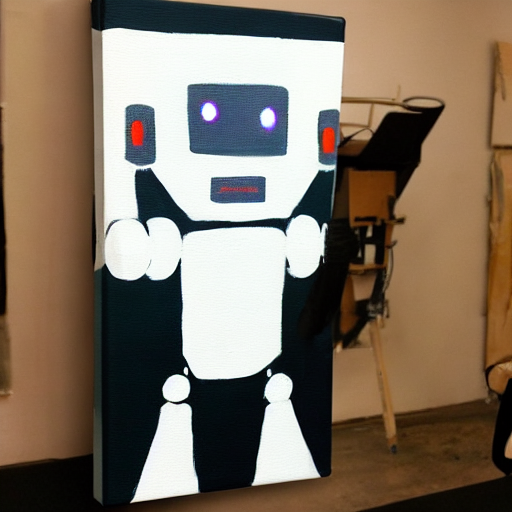} \end{tabular} \\
\hline
\end{tabular}
\caption{Prompt: \textit{Robot painting canvas in studio}}
\label{tab:Robot painting canvas in studio}
\end{table}

\begin{table}[h]
\centering
\begin{tabular}{|c|c|c|}
\hline
\textbf{Seed} & CONFORM  & SHYI  \\
\hline
4913 & \begin{tabular}{c} \tiny{USER: xx\%, xx\%} \\[-0.2cm] \tiny{TIFA: 1.0} \\[-0.2cm] \tiny{CLIP: 0.31} \\ \includegraphics[width=2.7cm]{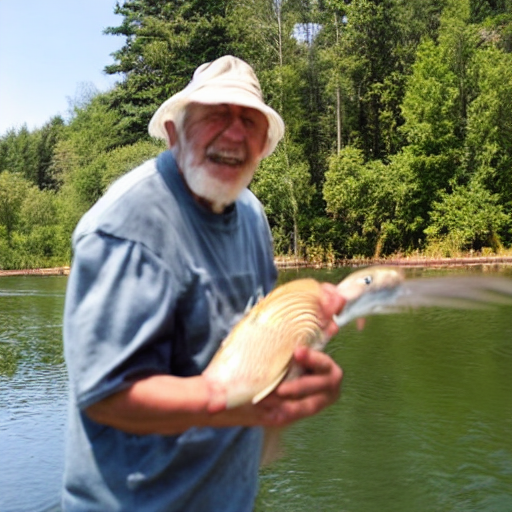} \end{tabular} & \begin{tabular}{c} \tiny{USER: xx\%, xx\%} \\[-0.2cm] \tiny{TIFA: 1.0} \\[-0.2cm] \tiny{CLIP: 0.31} \\ \includegraphics[width=2.7cm]{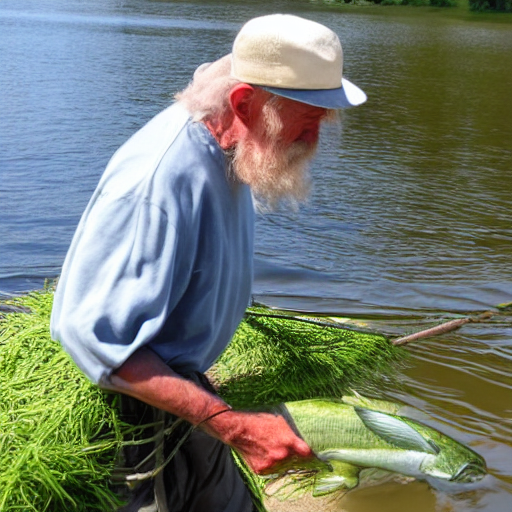} \end{tabular} \\
\hline
4914 & \begin{tabular}{c} \tiny{USER: xx\%, xx\%} \\[-0.2cm] \tiny{TIFA: 1.0} \\[-0.2cm] \tiny{CLIP: 0.27} \\ \includegraphics[width=2.7cm]{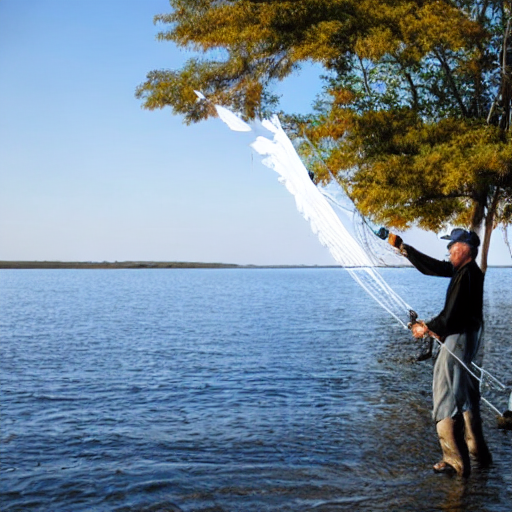} \end{tabular} & \begin{tabular}{c} \tiny{USER: xx\%, xx\%} \\[-0.2cm] \tiny{TIFA: 1.0} \\[-0.2cm] \tiny{CLIP: 0.31} \\ \includegraphics[width=2.7cm]{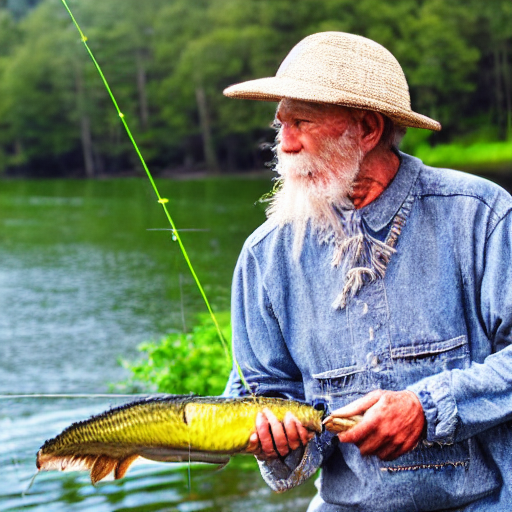} \end{tabular} \\
\hline
4915 & \begin{tabular}{c} \tiny{USER: xx\%, xx\%} \\[-0.2cm] \tiny{TIFA: 1.0} \\[-0.2cm] \tiny{CLIP: 0.31} \\ \includegraphics[width=2.7cm]{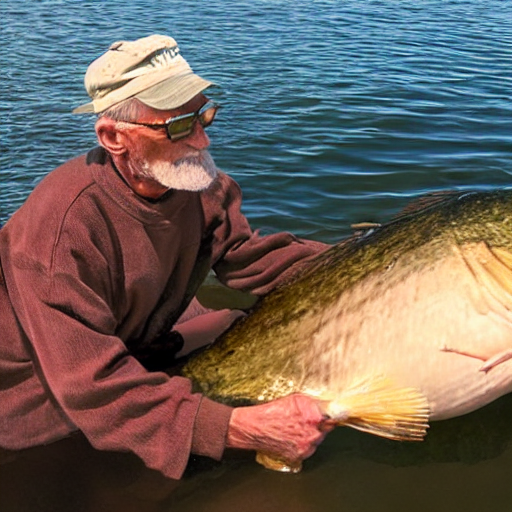} \end{tabular} & \begin{tabular}{c} \tiny{USER: xx\%, xx\%} \\[-0.2cm] \tiny{TIFA: 0.86} \\[-0.2cm] \tiny{CLIP: 0.30} \\ \includegraphics[width=2.7cm]{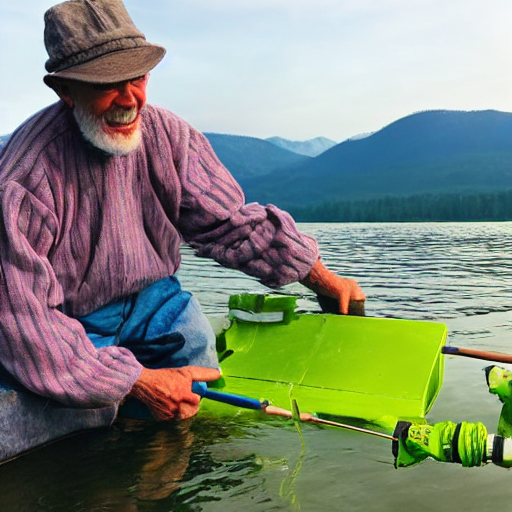} \end{tabular} \\
\hline
\end{tabular}
\caption{Prompt: \textit{Old fisherman reeling big fish near lake}}
\label{tab:Old fisherman reeling big fish near lake}
\end{table}

\begin{table}[h]
\centering
\begin{tabular}{|c|c|c|}
\hline
\textbf{Seed} & CONFORM  & SHYI  \\
\hline
4913 & \begin{tabular}{c} \tiny{USER: xx\%, xx\%} \\[-0.2cm] \tiny{TIFA: 1.0} \\[-0.2cm] \tiny{CLIP: 0.34} \\ \includegraphics[width=2.7cm]{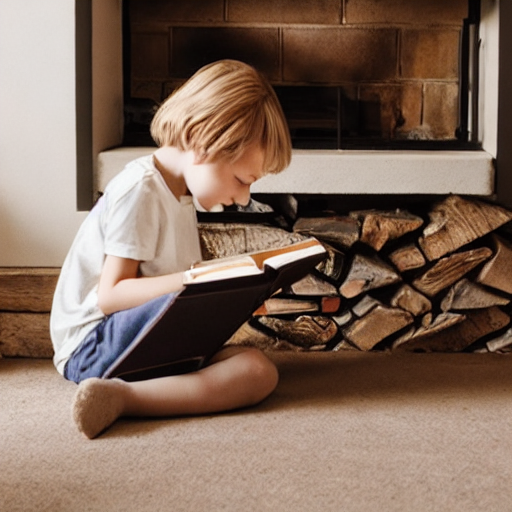} \end{tabular} & \begin{tabular}{c} \tiny{USER: xx\%, xx\%} \\[-0.2cm] \tiny{TIFA: 1.0} \\[-0.2cm] \tiny{CLIP: 0.33} \\ \includegraphics[width=2.7cm]{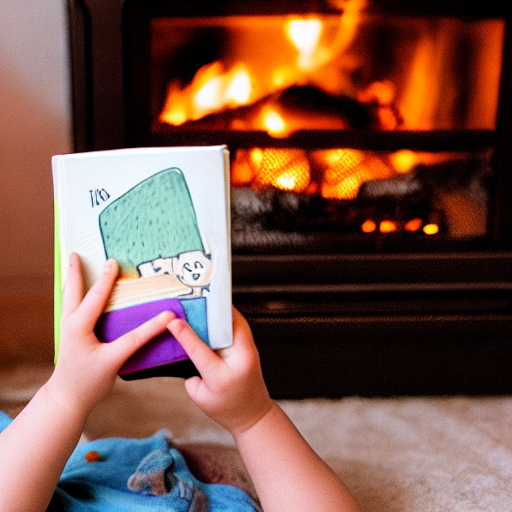} \end{tabular} \\
\hline
4914 & \begin{tabular}{c} \tiny{USER: xx\%, xx\%} \\[-0.2cm] \tiny{TIFA: 1.0} \\[-0.2cm] \tiny{CLIP: 0.35} \\ \includegraphics[width=2.7cm]{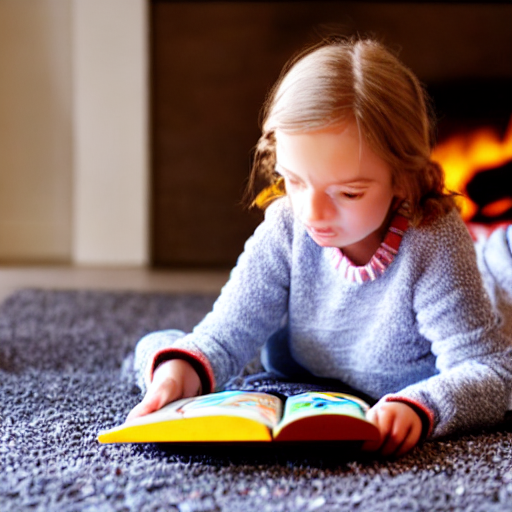} \end{tabular} & \begin{tabular}{c} \tiny{USER: xx\%, xx\%} \\[-0.2cm] \tiny{TIFA: 1.0} \\[-0.2cm] \tiny{CLIP: 0.31} \\ \includegraphics[width=2.7cm]{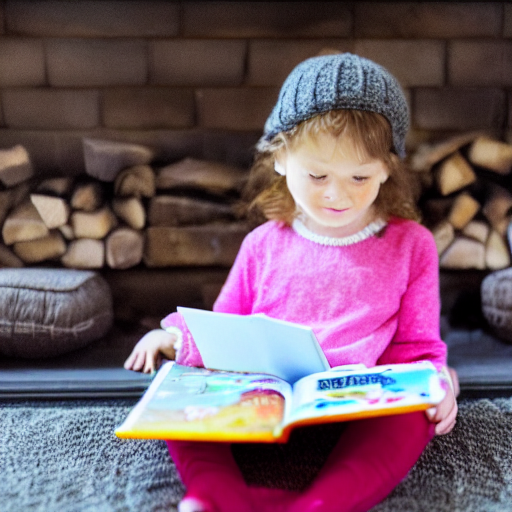} \end{tabular} \\
\hline
4915 & \begin{tabular}{c} \tiny{USER: xx\%, xx\%} \\[-0.2cm] \tiny{TIFA: 1.0} \\[-0.2cm] \tiny{CLIP: 0.35} \\ \includegraphics[width=2.7cm]{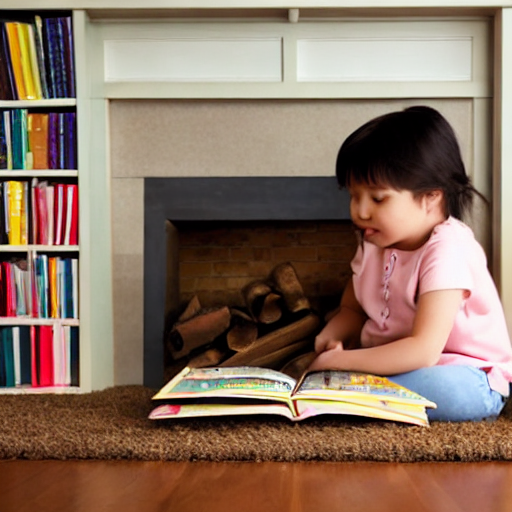} \end{tabular} & \begin{tabular}{c} \tiny{USER: xx\%, xx\%} \\[-0.2cm] \tiny{TIFA: 1.0} \\[-0.2cm] \tiny{CLIP: 0.34} \\ \includegraphics[width=2.7cm]{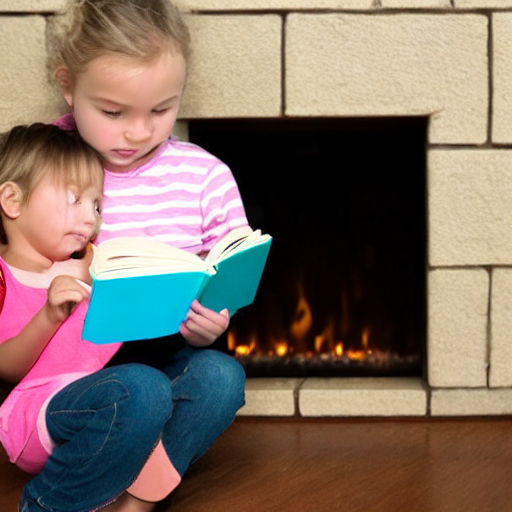} \end{tabular} \\
\hline
\end{tabular}
\caption{Prompt: \textit{Child reading book by fireplace}}
\label{tab:Child reading book by fireplace}
\end{table}

\begin{table}[h]
\centering
\begin{tabular}{|c|c|c|}
\hline
\textbf{Seed} & CONFORM  & SHYI  \\
\hline
4913 & \begin{tabular}{c} \tiny{USER: xx\%, xx\%} \\[-0.2cm] \tiny{TIFA: 0.83} \\[-0.2cm] \tiny{CLIP: 0.32} \\ \includegraphics[width=2.7cm]{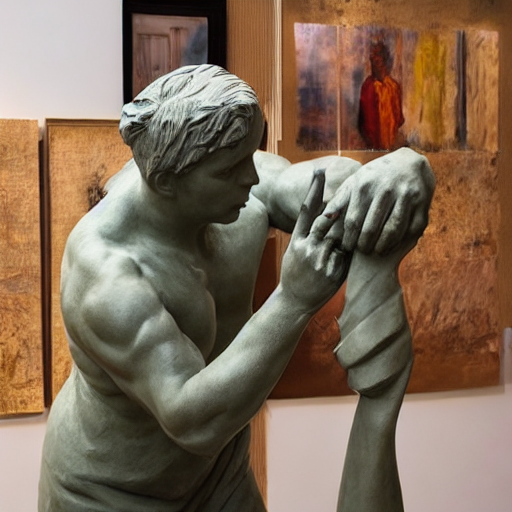} \end{tabular} & \begin{tabular}{c} \tiny{USER: xx\%, xx\%} \\[-0.2cm] \tiny{TIFA: 0.83} \\[-0.2cm] \tiny{CLIP: 0.33} \\ \includegraphics[width=2.7cm]{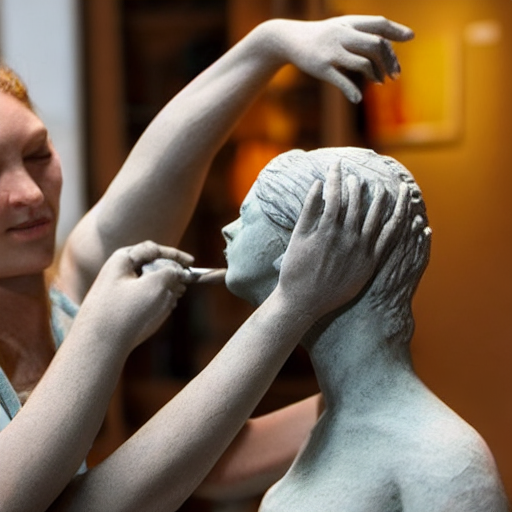} \end{tabular} \\
\hline
4914 & \begin{tabular}{c} \tiny{USER: xx\%, xx\%} \\[-0.2cm] \tiny{TIFA: 0.83} \\[-0.2cm] \tiny{CLIP: 0.30} \\ \includegraphics[width=2.7cm]{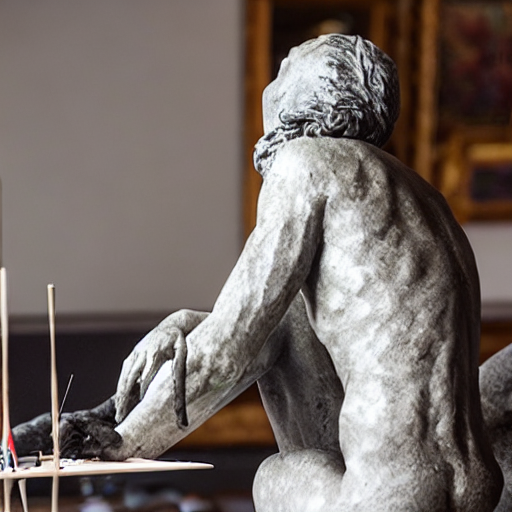} \end{tabular} & \begin{tabular}{c} \tiny{USER: xx\%, xx\%} \\[-0.2cm] \tiny{TIFA: 0.83} \\[-0.2cm] \tiny{CLIP: 0.32} \\ \includegraphics[width=2.7cm]{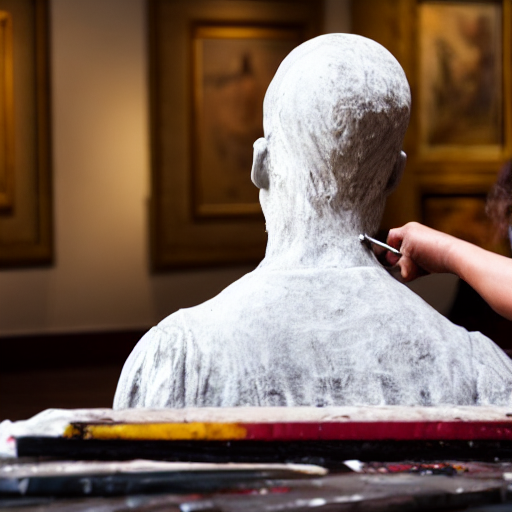} \end{tabular} \\
\hline
4915 & \begin{tabular}{c} \tiny{USER: xx\%, xx\%} \\[-0.2cm] \tiny{TIFA: 0.83} \\[-0.2cm] \tiny{CLIP: 0.30} \\ \includegraphics[width=2.7cm]{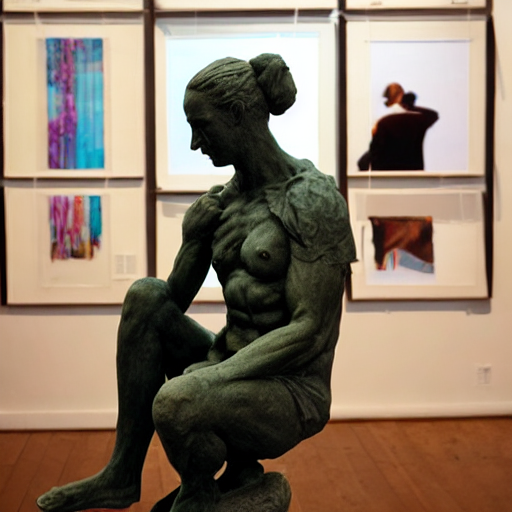} \end{tabular} & \begin{tabular}{c} \tiny{USER: xx\%, xx\%} \\[-0.2cm] \tiny{TIFA: 0.83} \\[-0.2cm] \tiny{CLIP: 0.31} \\ \includegraphics[width=2.7cm]{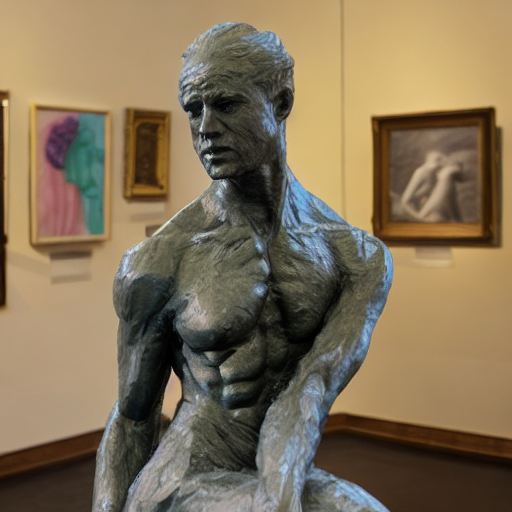} \end{tabular} \\
\hline
\end{tabular}
\caption{Prompt: \textit{Artist sculpting statue in gallery}}
\label{tab:Artist sculpting statue in gallery}
\end{table}

\begin{table}[h]
\centering
\begin{tabular}{|c|c|c|}
\hline
\textbf{Seed} & CONFORM  & SHYI  \\
\hline
4913 & \begin{tabular}{c} \tiny{USER: xx\%, xx\%} \\[-0.2cm] \tiny{TIFA: 0.89} \\[-0.2cm] \tiny{CLIP: 0.33} \\ \includegraphics[width=2.7cm]{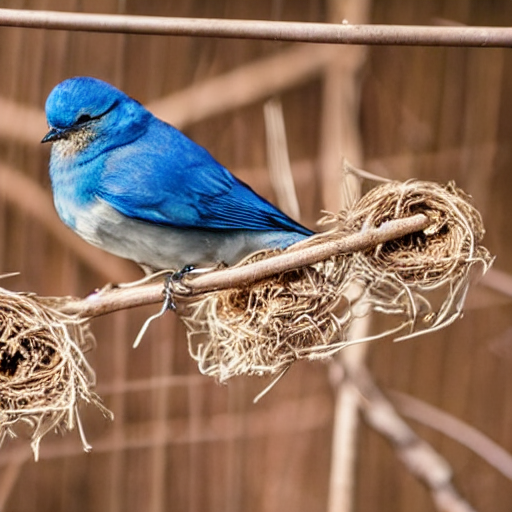} \end{tabular} & \begin{tabular}{c} \tiny{USER: xx\%, xx\%} \\[-0.2cm] \tiny{TIFA: 0.78} \\[-0.2cm] \tiny{CLIP: 0.30} \\ \includegraphics[width=2.7cm]{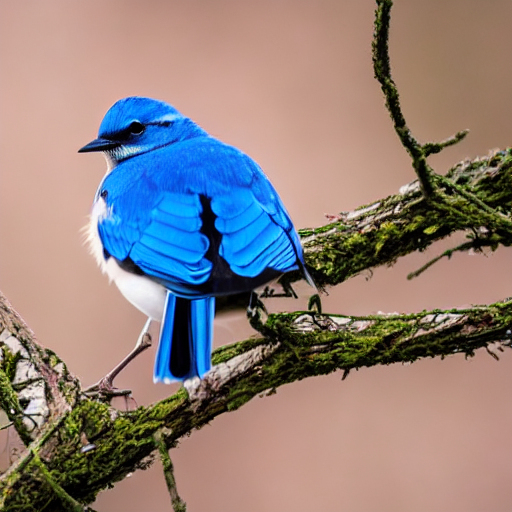} \end{tabular} \\
\hline
4914 & \begin{tabular}{c} \tiny{USER: xx\%, xx\%} \\[-0.2cm] \tiny{TIFA: 1.0} \\[-0.2cm] \tiny{CLIP: 0.32} \\ \includegraphics[width=2.7cm]{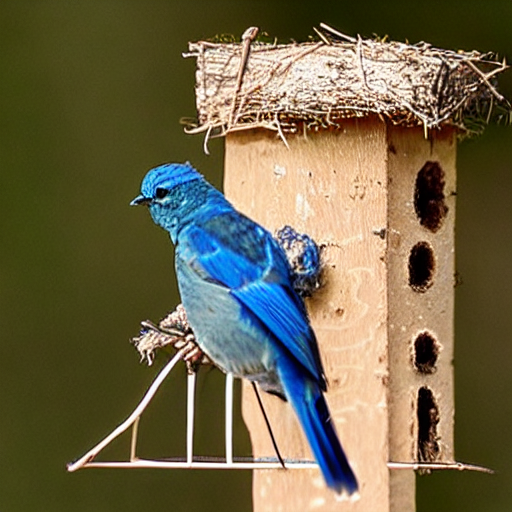} \end{tabular} & \begin{tabular}{c} \tiny{USER: xx\%, xx\%} \\[-0.2cm] \tiny{TIFA: 0.78} \\[-0.2cm] \tiny{CLIP: 0.31} \\ \includegraphics[width=2.7cm]{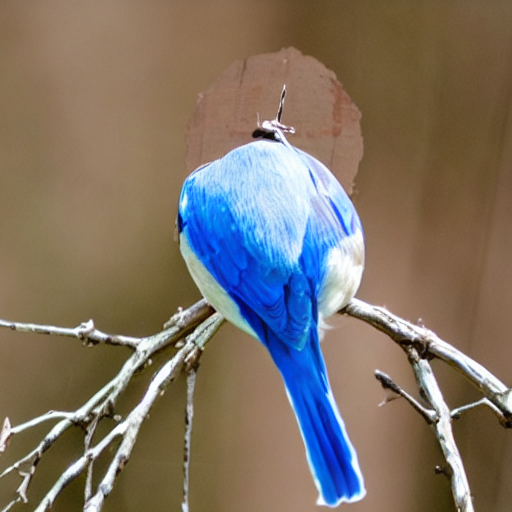} \end{tabular} \\
\hline
4915 & \begin{tabular}{c} \tiny{USER: xx\%, xx\%} \\[-0.2cm] \tiny{TIFA: 0.89} \\[-0.2cm] \tiny{CLIP: 0.32} \\ \includegraphics[width=2.7cm]{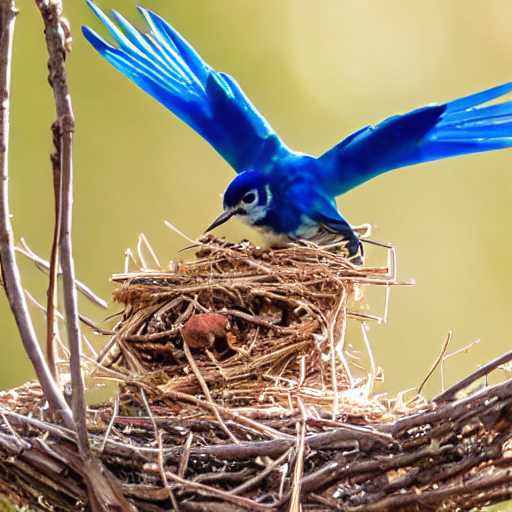} \end{tabular} & \begin{tabular}{c} \tiny{USER: xx\%, xx\%} \\[-0.2cm] \tiny{TIFA: 0.78} \\[-0.2cm] \tiny{CLIP: 0.28} \\ \includegraphics[width=2.7cm]{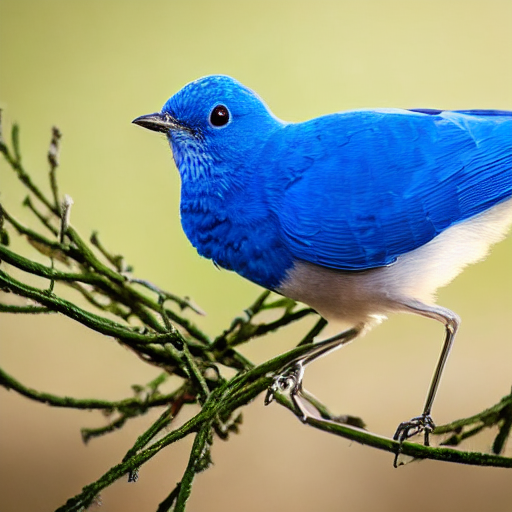} \end{tabular} \\
\hline
\end{tabular}
\caption{Prompt: \textit{Blue bird weaving tiny nest on branch}}
\label{tab:Blue bird weaving tiny nest on branch}
\end{table}

\begin{table}[h]
\centering
\begin{tabular}{|c|c|c|}
\hline
\textbf{Seed} & CONFORM  & SHYI  \\
\hline
4913 & \begin{tabular}{c} \tiny{USER: xx\%, xx\%} \\[-0.2cm] \tiny{TIFA: 0.71} \\[-0.2cm] \tiny{CLIP: 0.34} \\ \includegraphics[width=2.7cm]{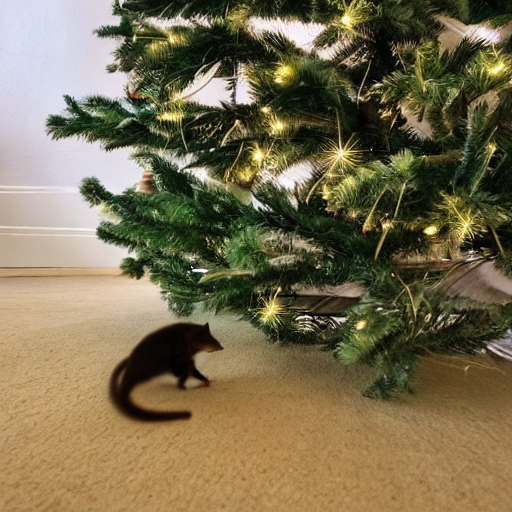} \end{tabular} & \begin{tabular}{c} \tiny{USER: xx\%, xx\%} \\[-0.2cm] \tiny{TIFA: 0.71} \\[-0.2cm] \tiny{CLIP: 0.31} \\ \includegraphics[width=2.7cm]{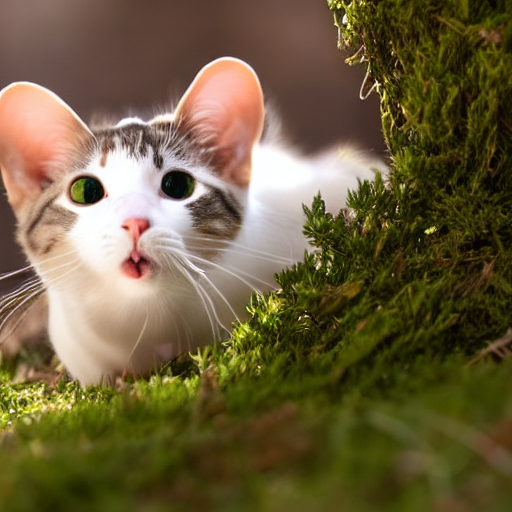} \end{tabular} \\
\hline
4914 & \begin{tabular}{c} \tiny{USER: xx\%, xx\%} \\[-0.2cm] \tiny{TIFA: 1.0} \\[-0.2cm] \tiny{CLIP: 0.32} \\ \includegraphics[width=2.7cm]{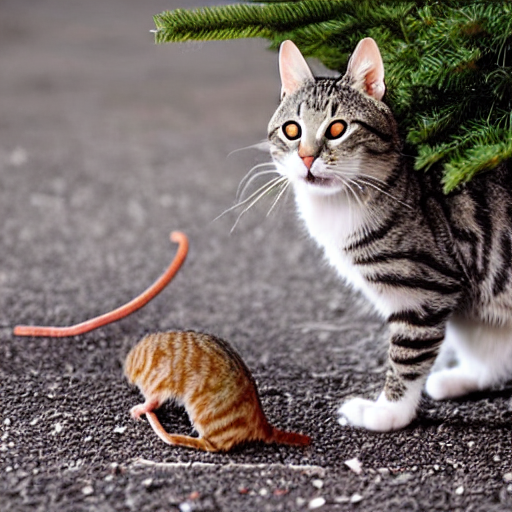} \end{tabular} & \begin{tabular}{c} \tiny{USER: xx\%, xx\%} \\[-0.2cm] \tiny{TIFA: 0.71} \\[-0.2cm] \tiny{CLIP: 0.28} \\ \includegraphics[width=2.7cm]{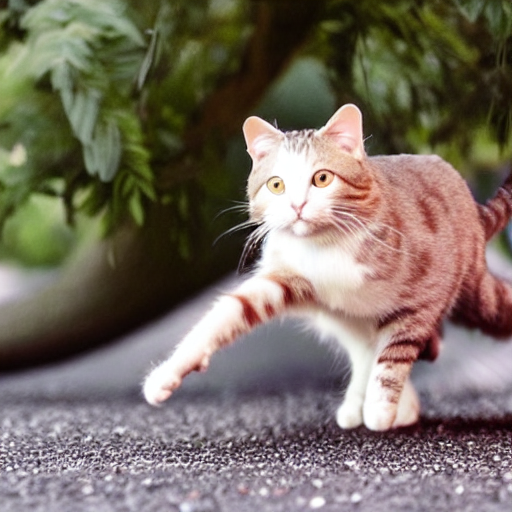} \end{tabular} \\
\hline
4915 & \begin{tabular}{c} \tiny{USER: xx\%, xx\%} \\[-0.2cm] \tiny{TIFA: 0.71} \\[-0.2cm] \tiny{CLIP: 0.32} \\ \includegraphics[width=2.7cm]{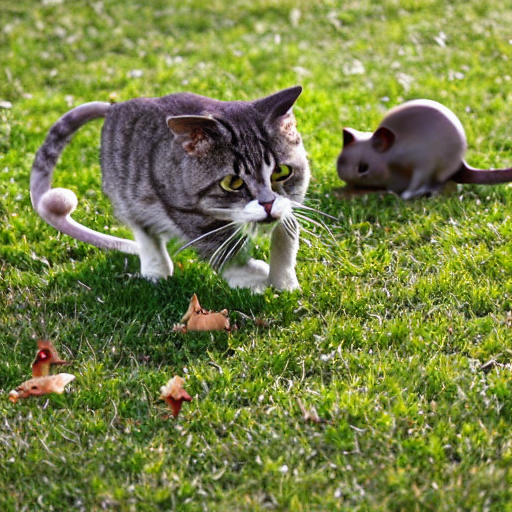} \end{tabular} & \begin{tabular}{c} \tiny{USER: xx\%, xx\%} \\[-0.2cm] \tiny{TIFA: 0.57} \\[-0.2cm] \tiny{CLIP: 0.31} \\ \includegraphics[width=2.7cm]{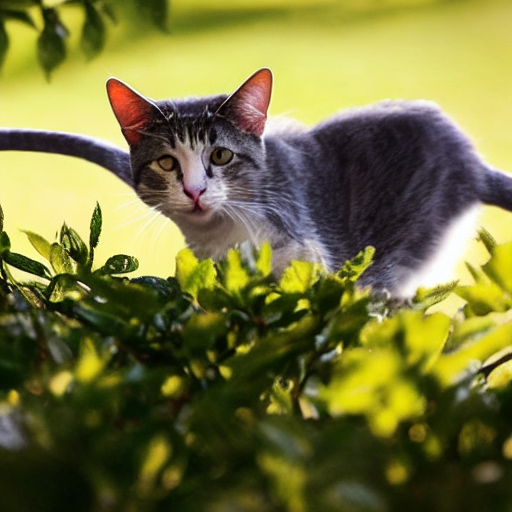} \end{tabular} \\
\hline
\end{tabular}
\caption{Prompt: \textit{Cat chasing mouse under tree}}
\label{tab:Cat chasing mouse under tree}
\end{table}

\end{document}